\documentclass{article}
\PassOptionsToPackage{numbers,sort&compress}{natbib}
\usepackage[preprint]{neurips_2026}
\usepackage{graphicx}
\usepackage{booktabs}
\usepackage{amssymb}

\newcommand{\eg}{\textit{e.g.}}

\usepackage[table]{xcolor}
\usepackage{soul}
\definecolor{tabfirst}{rgb}{1, 0.7, 0.7} % red
\definecolor{tabsecond}{rgb}{1, 0.85, 0.7} % orange
\definecolor{tabthird}{rgb}{1, 1, 0.7} % yellow
\usepackage{amsmath}
\usepackage{caption}
\usepackage{wrapfig}
\usepackage{marvosym}

\author{%
  \textbf{Linrui Dai}\textsuperscript{1, 2} \quad
  \textbf{Jinxiu Liang}\textsuperscript{2}\thanks{Corresponding author.} \quad
  \textbf{Chu Zhou}\textsuperscript{2} \\
  \textbf{Mae Yamaguchi\textsuperscript{1, 2}} \quad
  \textbf{Ryuya Takahashi\textsuperscript{1, 2}} \quad
  \textbf{Imari Sato\textsuperscript{1, 2}} \\[4pt]
  \textsuperscript{1}The University of Tokyo \\
  \textsuperscript{2}National Institute of Informatics \\
}

\usepackage[pagebackref,breaklinks,colorlinks]{hyperref}
\usepackage{multicol}
\usepackage{multirow,booktabs}
\usepackage{array}
\usepackage{setspace,tabularx,graphicx}
\usepackage{subcaption}
\usepackage{algorithm}
\usepackage{algpseudocode}

\usepackage{orcidlink}

\begin{document}

\title{Extremely Sparse-View Computed Tomography from 2D Projections via Pose-Aware Diffusion Priors}

\maketitle

\begin{abstract}
Extremely sparse-view computed tomography (xSV-CT) is critical when scanning time, motion, or radiation tolerance limits each scan to only a few projections. In many such scenarios, the same acquisition limits also constrain dataset construction, making pre-reconstructed 3D volumes from dense scans unavailable. This exposes a gap in current sparse-view approaches: Optimization-based methods, which use the sparse measurements directly but lack data-driven priors to suppress artifacts, and learning-based methods, which use strong priors but usually require densely scanned training volumes acquired outside the xSV-CT regime. We study this train-sparse / infer-sparse setting by learning a pose-aware diffusion prior from one 2D projection per training sample, then guiding diffusion sampling with measured-view fidelity, shared-volume reprojection, and projection-physics consistency for tomographic reconstruction. On two public CT datasets across 1-, 4-, and 8-view settings, the proposed method achieves the best novel-view synthesis scores and best or competitive reconstruction quality compared with baselines that use dense-volume supervision.
\end{abstract}

\vspace{-14pt}

\section{Introduction}
\label{sec:intro}
Computed tomography (CT) is essential for 3D diagnosis, but dense angular sampling is time-consuming and radiation-intensive. While sparse-view CT mitigates these problems by reconstructing from a sparser set of projections~\cite{di2023review,lin2023learning,lin2024learning,zhang2025x}, several applications demand more extreme reduction, allowing only a few snapshots. This includes intra-operative C-arm navigation at 3-8 views~\cite{otomo2022computed}, temporally dynamic anatomy imaging such as cardiac CT~\cite{schondube2013temporal}, and radiation-sensitive specimens that reach their absorption limit within one or two scans~\cite{garman2021radiation}. These scenarios motivate an extremely sparse-view CT (xSV-CT) regime: high-quality reconstruction from fewer than 10, or even a single, projection. Unlike prior formulations that aim to reduce radiation footprint at reconstruction~\cite{lin2023learning,lin2024learning,zhang2025x}, these applications impose acquisition constraints that extend beyond test-time reconstruction: The same acquisition limits can also make densely sampled projections or pre-reconstructed 3D training volumes difficult to collect for the target regime on a large scale for model training.

\begin{figure}
% \vspace{-10pt}
\includegraphics[width=0.98\linewidth]{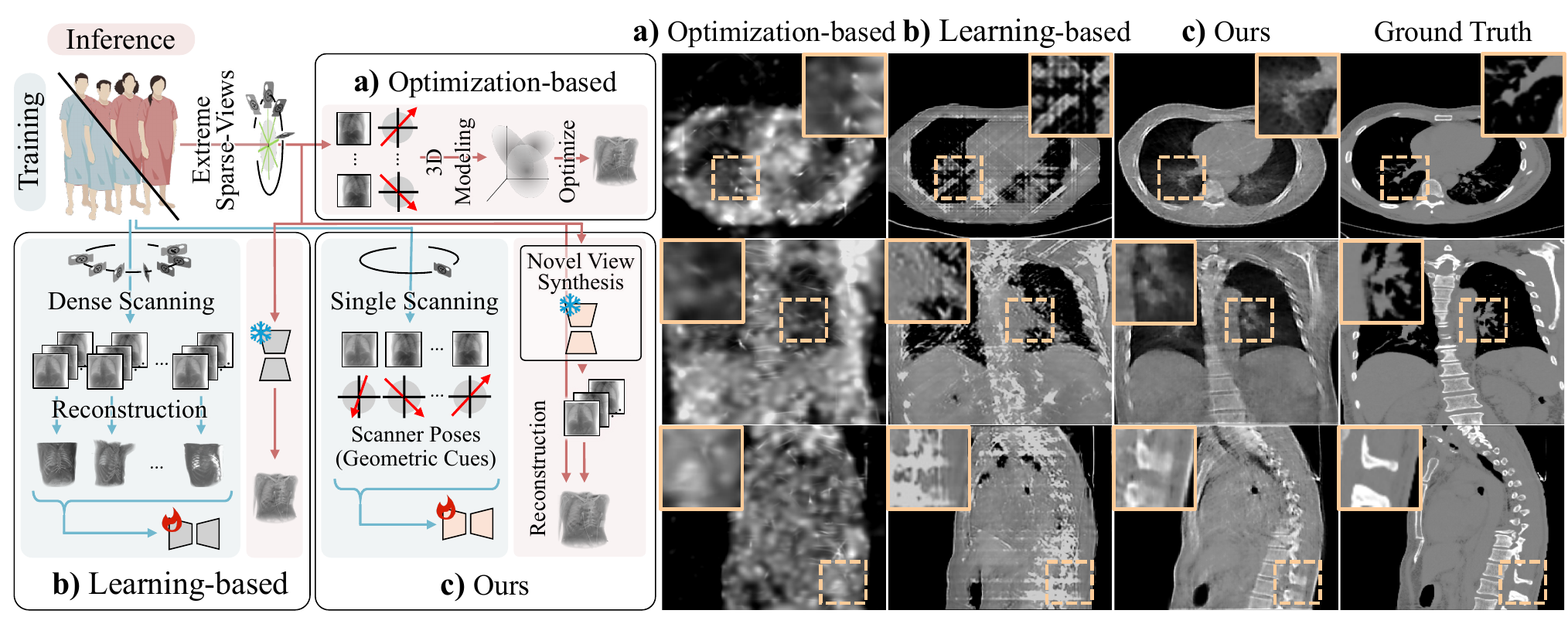}
\captionof{figure}{Tomographic reconstruction results from extremely sparse views at $0^\circ$, $45^\circ$, $90^\circ$, $135^\circ$. Current methods cover complementary sides of the extremely sparse view reconstruction problem: \textbf{a}) optimization-based methods~\cite{cai2024radiative} directly reconstruct from input views but produces severe artifacts when pushed to extreme sparsity at inference. \textbf{b}) Learning-based methods benefit from data-driven priors, but typically require reconstructed datasets from dense scannings that are difficult to obtain for certain applications. \textbf{c}) To simultaneously respond to the sparsity limit constraining both the training and inference process, our method learns a pose-aware projection-domain diffusion prior \textit{from a single 2D projection per training sample}, then uses physics-guided sampling to produce tomographically consistent novel projections for accurate reconstruction.}  
\label{fig:teaser} 
\vspace{-20pt}
\end{figure}

Existing SV-CT methods cover complementary sides of this acquisition problem. \textit{Optimization-based methods} recover a 3D representation per sample using only its sparse input projections as supervision~\cite{zha2022naf,cai2024structure,zha2024r,lin2024learning,cai2024radiative,li20253dgr}. This formulation automatically admits the xSV-CT regime. However, when pushed to extreme sparsity, their lack of anatomical priors to constrain the solution space results in severe reconstruction artifacts (Fig.~\ref{fig:teaser}(\textbf{a})). 
\textit{Learning-based methods} learn strong statistical priors from large-scale datasets of pre-reconstructed 3D volumes~\cite{lin2023learning,lin2024c,zhang2025x,pan2025draw,xie2025dvg,liu2025d3t} and have demonstrated anatomically coherent reconstructions from 4-10 views. Their training signal usually comes from data acquired outside the target xSV-CT regime: each training volume or slice assumes sufficient angular coverage for reliable reconstruction (Fig.~\ref{fig:teaser}(\textbf{b})). For fragile, dynamic, or dose-sensitive targets, this training-data requirement becomes the main bottleneck.

\textit{Can we reduce X-ray scans at both training and inference?} This setting prompts us to change the learning signal. A single training projection does not identify an individual 3D volume, but across many objects, projection--pose pairs define a pose-conditioned distribution of physically plausible measurements. If samples from this distribution are constrained at inference to agree with measured views and tomographic consistency, the learned projection prior can supply missing angular information without dense training scans. Motivated by this observation, we train a diffusion model on pairs of a projection and its acquisition angle, accessing \textit{only one 2D projection per training sample}. At inference, we use this model to synthesize a dense set of candidate projections from sparse-view inputs for standard reconstruction (Fig.~\ref{fig:teaser}(\textbf{c})).

\begin{figure*}[t]
    \centering
    \includegraphics[width=\linewidth]{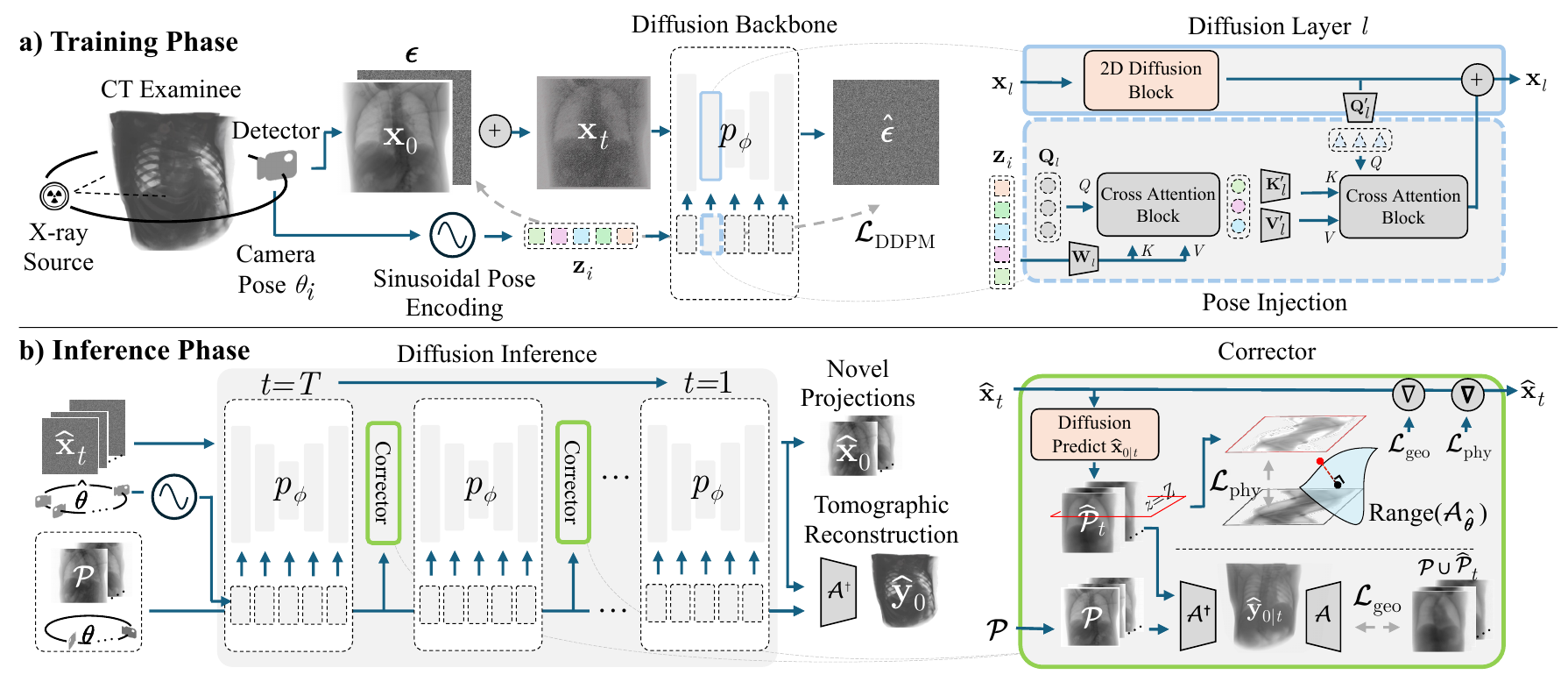}
    \caption{Overview of our framework. \textbf{a}) Training phase (Sec.~\ref{sec:3.1}): we learn a \textbf{pose-aware projection-domain diffusion prior} from 2D projections $\mathbf{x}_0$ paired with poses $\theta_i$ via network $p_\phi$. Specifically, we prepare geometric cues $\theta_i$ via \textit{sinusoidal pose encoding} and inject them with \textit{multi-scale pose injection} (right). The model is trained with standard denoising loss $\mathcal{L}_{\rm DDPM}$. \textbf{b}) Inference phase (Sec.~\ref{sec:3.2}): given $M$ sparse input views $\mathcal{P}$ at poses $\boldsymbol{\theta}$, we generate $N$ novel views $\hat{\mathbf{x}}_0$ at poses $\hat{\boldsymbol{\theta}}$ through \textbf{consistency-aware guided generation}. At each denoising timestep $t$, a \textit{corrector module} (right) enforces two complementary constraints: \textit{geometric coherency} loss $\mathcal{L}_{\rm geo}$ pushes all generated and input views toward a shared 3D explanation via forward projection $\mathcal{A}$ and approximate inverse $\mathcal{A}^\dagger$; \textit{physics-informed consistency} loss $\mathcal{L}_{\rm phy}$ further enhances consistency among generated views $\hat{\mathcal{P}}_t$ by projecting them onto the consistent range of the Radon transform $\mathcal{A}_{\hat{\boldsymbol{\theta}}}$.}
    \label{fig:mainfig}\vspace{-10pt}
\end{figure*}

Following this strategy, we separate the problem into a learning component and an inverse-problem component. (\textbf{1}) For training (Fig. \ref{fig:mainfig}(\textbf{a})), we leverage the strong distribution modeling capability from diffusion models widely used in CT~\cite{dai2026guidegen} and propose a \emph{pose-aware projection-domain diffusion model} that injects scanner-angle geometry via sinusoidal pose encoding and multi-scale cross-attention at every network depth. This lets a 2D diffusion backbone model pose-conditioned projection statistics, including the reflection symmetry between complementary parallel-beam angles and the local sensitivity needed for adjacent views. (\textbf{2}) For inference (Fig.~\ref{fig:mainfig}(\textbf{b})), we introduce \emph{consistency-aware guided generation} that steers the reverse diffusion process with two constraints: a \emph{geometric coherency} loss that ties generated and measured projections to a shared reconstructed volume, and a \emph{physics-informed consistency} loss that enforces consistency among generated views by projecting them onto a valid sinogram.

We evaluate our formulation on two public CT datasets with projection-volume pairs simulated under conventional CT noise models or from a verified forward model CATSim~\cite{catsim}. Reconstruction volumes are held out from our method and used only as metric references. Across 1-, 4-, and 8-view settings, our method achieves the best novel-view synthesis scores and competitive reconstruction quality against learning-based methods trained on full 3D reconstructions. This shows that, unlike uncorrected or feed-forward generated projections~\cite{su2026diffnr}, a guided projection-domain sampler optimized with appropriate signals could also provide valuable information for sparse-view CT reconstruction.

Our main contributions are threefold: (\textbf{1}) We study xSV-CT under a train-sparse / infer-sparse setting with arbitrary acquisition poses, learning a pose-aware projection-domain diffusion prior from a single 2D projection per training sample and thereby removing the need for dense-volume supervision in the target data regime; (\textbf{2}) We integrate sinusoidal pose encoding with multi-scale cross-attention injection, enabling a 2D diffusion backbone to model pose-conditioned projection distributions from per-sample projections; (\textbf{3}) We embed a discrete physical surrogate directly inside diffusion sampling, coupled with a shared-volume geometric coherency loss, so generated dense projections are refined toward a common tomographic explanation.

\section{Related Works}
\label{sec:2}

\noindent\textbf{Optimization-based methods} reconstruct each test case from its own sparse input projections, with no learned data prior from a training database.
Classical analytical methods such as filtered back-projection (FBP) accurately reconstruct from dense angular coverage, but severely degrade under sparse sampling due to the Fourier Slice Theorem: insufficient angular coverage leaves the inverse problem fundamentally ill-posed~\cite{anastasio2001comments,kudo2013image,kak2001principles}.
Iterative methods tackle sparse view tomographic reconstruction by minimizing energy functions with regularization~\cite{feldkamp1984practical,scales1987tomographic,wang2004ordered,sidky2008image}, often combining network-based~\cite{anirudh2018lose,zang2021intratomo} or plug-and-play denoisers to enforce smoothness~\cite{venkatakrishnan2013plug,zhang2017learning}.
Modern learning-based methods further adapt NeRF-based implicit representations~\cite{zha2022naf,cai2024structure} or 3D Gaussian splatting~\cite{zha2024r,lin2024learning,cai2024radiative,li20253dgr,wang2025xfield,wu2025discretized} to the CT geometry, achieving strong results at 20--50 views.
At fewer than 10 views, geometry alone leaves too much angular information missing, and artifacts become severe~\cite{xie2024prior}. This motivates a learned projection-domain prior that can be trained from 2D measurements while preserving the projection-domain inference setting.

% \subsection{Volume-based tomographic reconstruction}
\noindent\textbf{Learning-based methods} learn strong statistical priors over 3D anatomy from large-scale datasets of pre-reconstructed CT volumes~\cite{lin2023learning,lee2023improving,lin2024c,pan2025draw,zhang2025x,xie2025dvg,liu2025d3t}.
Volume-based regression methods learn a deterministic mapping from sparse projections to 3D volumes via deep unrolling~\cite{li2018optimization,zhang2019jsr}, GANs~\cite{pan2025draw}, or implicit neural fields~\cite{lin2023learning,lin2024learning,lin2024c}, supervised by ground-truth volumes obtained from dense-angle scans.
Volume-based diffusion methods train 2D prior models on anatomical slices extracted from CT volumes~\cite{chung2023solving,lee2023improving,he2024solving,xie2025dvg} and apply posterior inference to reconstruct slices conditioned on sparse measurements. Subsequent work has then substantially reduced the inference cost of such priors for large volumes~\cite{chung2024decomposed,song2024diffusionblend,du2026improving}.
These methods are computationally efficient at inference, but their training priors are learned from volumes reconstructed with dense angular coverage. For radiation-sensitive or rapidly changing targets, this creates a mismatch between the training-data regime and the acquisition regime of interest. Our approach learns a pose-aware projection-domain diffusion prior directly from 2D projections, removing dense-volume supervision during training while keeping measured-view fidelity and tomographic consistency at inference.

\textbf{Few-view X-ray-to-CT generation} methods have shown that networks can infer volumetric structure from one or a few radiographs at predefined poses, using deep regression~\cite{shen2019patient}, adversarial training~\cite{ying2019x2ct,zhao2024construction}, or diffusion-based multi-modal generation~\cite{zhan2024medm2g}. Like the learning-based methods above, they form their training pairs from reconstructed volumes and are designed around a fixed acquisition protocol, so changing the number or poses of input views typically requires redesign or retraining. Closer to our setting, Lai. \textit{et al.}~\cite{lai2024reconstructing} learn solely in the projection domain with a self-supervised neural field, but require 90 projections per training sample and a fixed biplanar protocol at inference. In contrast, we train on a single projection per sample and accept measurements at arbitrary poses, which admits a train-sparse setting and broadens potential downstream applications.

\section{Method}
The tomographic reconstruction problem can be formulated as finding the optimal solution to inverse the forward transform operator (\eg, Radon) $\mathcal{A}_{\boldsymbol{\theta}}$~\cite{chung2023solving}, given the projections $\mathbf{x}=\{\mathbf{x}_i\}$ from a set of input projection angles $\boldsymbol{\theta}=\{\theta_i\}$:
\begin{equation}
    \mathbf{x}_i=\mathcal{A}_{\theta_i}\mathbf{y}+\mathbf{n}_i,
\end{equation}
where $\mathbf{y}$ is the 3D volume and $\mathbf{n}_i$ is the noise tensor induced by photon scattering and electronic noise~\cite{yu2012development}. To achieve high-quality xSV-CT from extremely sparse sets of projections, we propose a framework that generates consistent projections for reconstruction via a projection-domain generative prior trained from a single 2D projection per training sample (Fig.~\ref{fig:mainfig}). The prior models angle-conditioned projection statistics rather than identifying a 3D volume from one training view. At inference, measured-view fidelity and tomographic consistency turn this prior into a constrained reconstruction procedure (Sec.~\ref{sec:3.2}).

\subsection{Pose-Aware Projection-Domain Diffusion Prior}
\label{sec:3.1}
We model the underlying data distribution of 2D X-ray projections using a denoising diffusion probabilistic model (DDPM)~\cite{ho2020denoising}. To generate projections that are consistent with different scanner angles at inference, we condition the model on the scanner pose at every layer of the network. Formally, during training the diffusion forward process progressively corrupts a clean projection $\mathbf{x}_0$ by adding Gaussian noise $\boldsymbol{\epsilon}$ over $T$ timesteps via a variance schedule $\beta_t$ ($0 \leq t \leq T$). At timestep $t$, the noisy projection $\mathbf{x}_t$ is generated as:
\begin{equation}
    \mathbf{x}_t=\sqrt{\bar{\alpha}_t}\mathbf{x}_0+\sqrt{1-\bar{\alpha}_t}\boldsymbol{\epsilon},
\end{equation}
where $\bar{\alpha}_t=\prod_{\tau=1}^t\alpha_\tau$ and $\alpha_t=1-\beta_t$. The diffusion model $p_\phi$ is then trained to reverse this degradation by predicting the added noise $\boldsymbol{\epsilon}$ from $\mathbf{x}_t$ via:
\begin{equation}
    \mathcal{L}_{\rm DDPM}=\mathbb{E}_{t,\mathbf{x}_0,\boldsymbol{\epsilon}}||p_\phi(\mathbf{x}_t;\theta_i,t)-\boldsymbol{\epsilon}||_2^2,
\end{equation}
where $\theta_i$ is the detector pose associated with the input projection $\mathbf{x}_0$. We promote the pose-awareness of our model $p_\phi$ through the following two components.

\begin{figure}[t]
    \centering
    \includegraphics[width=0.8\linewidth]{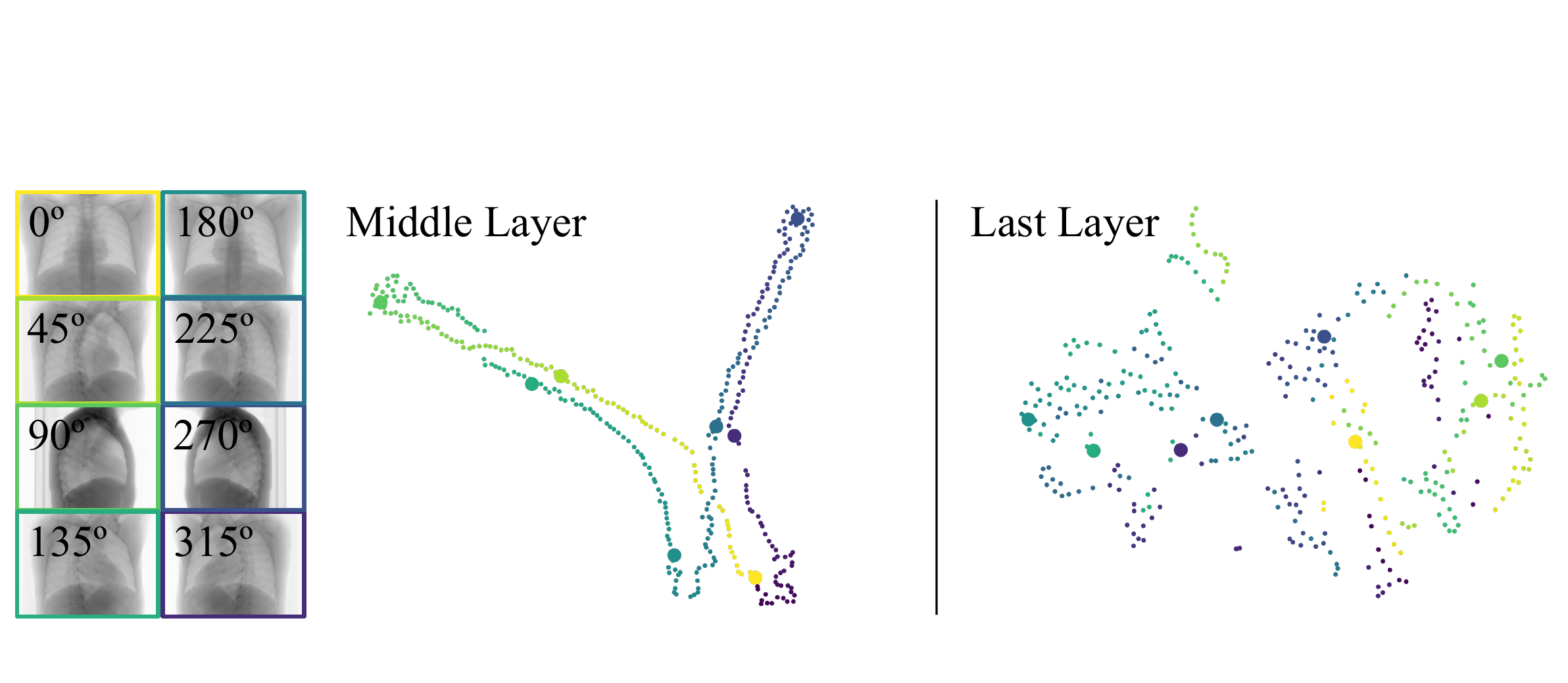}\vspace{-10pt}
    \caption{t-SNE visualizations~\cite{maaten2008visualizing} of learned features $\mathbf{z}_l$ at different layers. Larger dots in the feature space highlight features from the displayed sample projections. $\mathbf{z}_l$ at the middle layer (left) aligns features with the general structural distinctiveness of projections to resolve view ambiguity, while $\mathbf{z}_l$ at the last layer (right) retains high local sensitivity to individual angles, preserving a separable feature space for accurate pose-conditioned synthesis.}
    \label{fig:latent}
    \vspace{-10pt}
\end{figure}

\noindent\textbf{Sinusoidal Pose Encoding}
Due to X-ray transmissivity, projections from complementary angles can be closely related (\eg, projections at $0^\circ$ and $180^\circ$ in Fig.~\ref{fig:latent} under parallel-beam geometry differ mainly by detector reflection), while adjacent angles can contain small but important changes. 
We encode each scalar angle $\theta_i$ (pose) into a $K$-dimensional feature vector $\mathbf{z}_i$ with a sinusoidal encoding module:
\begin{equation}
    \mathbf{z}_i[k]=\sin\!\left((\lfloor k/2\rfloor+1)\,\theta_i+|k\pi/2|\right).
\end{equation}
The periodic basis gives the network a smooth angular coordinate system. It does not assume that complementary views are identical; instead, it exposes the symmetry and local angular variation in a form that cross-attention layers can learn from data.
% For even $k=2m$: $\mathbf{z}_i[2m]=\sin((m+1)\theta_i)$; for odd $k=2m+1$: $\mathbf{z}_i[2m+1]=\cos((m+1)\theta_i)$. The two components $(k{=}2m,\,k{=}2m+1)$ form a pair $(\sin((m+1)\theta),\cos((m+1)\theta))$, so $\|\mathbf{z}_i[2m{:}2m+1]\|=1$ for all $\theta_i$. Since $\sin((m+1)(\theta+\pi))=(-1)^{m+1}\sin((m+1)\theta)$ and $\cos((m+1)(\theta+\pi))=(-1)^{m+1}\cos((m+1)\theta)$, supplementary angles share the same representation magnitude for all $m$, capturing the physical transmissivity symmetry. 
High-frequency components ($k$ large) provide fine-grained angular sensitivity for detailed projection synthesis.

\noindent\textbf{Multi-Scale Pose Injection} 
Given encoded pose features $\mathbf{z}_i\in\mathbb{R}^K$ for $\theta_i$, we inject the pose information throughout the diffusion backbone via layer-wise cross-attention mechanism~\cite{vaswani2017attention}. At each layer $l$, we compute a layer-specific pose feature $\mathbf{z}_l$ through:
\begin{equation}
    \mathbf{z}_l={\texttt{CrossAttn}}\left(\mathbf{Q}_l,\mathbf{W}_l(\mathbf{z}_i),\mathbf{W}_l(\mathbf{z}_i)\right),
\end{equation}
(See Fig.\ref{fig:mainfig}(\textbf{a}), right), where  $\mathbf{Q}_l$ is a learnable query and $\mathbf{W}_l$ is a learnable projection. The cross-attention operation is defined as
$\texttt{CrossAttn}(Q,K,V)=\texttt{MLP}(\texttt{LayerNorm}(\texttt{Softmax}(\frac{QK^\top}{\sqrt{d}})V))$
with $d$ being the size of the latent dimension of $K$~\cite{vaswani2017attention}. This allows each layer to learn which components of the pose information $\mathbf{z}_i$ are the most relevant at its feature scale. As shown in Fig.~\ref{fig:latent}, layers with larger receptive fields (middle layers) tend to collapse pose features based on the general structural distinctiveness of projections to resolve view ambiguity, while layers with smaller receptive fields (later layers) tend to preserve the fine-grained separability of adjacent views on the feature space for detailed synthesis. The pose feature $\mathbf{z}_l$ is then fused with the intermediate image feature map $\mathbf{x}_l$ via a second cross-attention module and a residual connection:
\begin{equation}
    \mathbf{h}_l\leftarrow\texttt{CrossAttn}\!\left(\mathbf{Q}'_l(\mathbf{h}_l),\,\mathbf{K}'_l(\mathbf{z}_l),\,\mathbf{V}'_l(\mathbf{z}_l)\right)+\mathbf{h}_l,
\end{equation}
where $\mathbf{Q}'_l,\mathbf{K}'_l,\mathbf{V}'_l$ are learnable projections mapping to a common latent dimension. %The standard cross-attention here treats $\mathbf{z}_l$ as a single-token context sequence, so each spatial position of $\mathbf{h}_l$ attends to the pose embedding and updates accordingly. This multi-scale injection lets the model attend to both population-level structural variability and subtle, pose-dependent cues, learning features that implicitly encode how anatomy projects differently across viewpoints.

\subsection{Consistency-Aware Guided Generation}
\label{sec:3.2}

After training the pose-aware projection-domain diffusion model $p_\phi$, we generate a dense set of $N$ novel projections $\hat{\mathbf{x}}_0$ at poses $\hat{\boldsymbol{\theta}}$ that are geometrically coherent with each other and consistent with the $M$ sparse input measurements $\mathcal{P}$ at poses $\boldsymbol{\theta}$. Standard reverse sampling in DDPM produces clean $\hat{\mathbf{x}}_0$ from $p_\phi$ via repeated denoising from Gaussian noise~\cite{ho2020denoising}:
\begin{equation}
\label{ddpminference}
    \hat{\mathbf{x}}_{t-1}=\frac{1}{\sqrt{\alpha_t}}\left(\hat{\mathbf{x}}_t-\frac{\beta_t}{\sqrt{1-\bar{\alpha}_t}}p_\phi(\hat{\mathbf{x}}_t;\theta,t)\right)+\beta_t\boldsymbol{\epsilon}.
\end{equation}
For xSV-CT, batch-wise independently generated projections must be coupled to the measured views and to a common projection geometry. Inspired by classifier guidance~\cite{dhariwal2021diffusion}, we guide the reverse diffusion process with geometry- and physics-based constraints.

At each denoising timestep $t$ and novel pose $\hat{\theta}\in\hat{\boldsymbol{\theta}}$, we predict the clean projection $\hat{\mathbf{x}}_{0|t}$ through the standard denoising update~\cite{ho2020denoising}:
\begin{equation}
    \hat{\mathbf{x}}_{0|t}=(\mathbf{x}_t-\sqrt{1-\bar{\alpha}_t}p_\phi(\mathbf{x}_t;\hat{\theta},t))/\sqrt{\bar{\alpha}_t} \in\hat{\mathcal{P}}_t,
\end{equation}
Let $\hat{\mathcal{P}}_t$ denote the collection of $\hat{\mathbf{x}}_{0|t}$ evaluated at all novel projection angles $\hat{\boldsymbol{\theta}}$. We enforce consistency via two complementary constraints.

\noindent\textbf{Geometric Coherency Loss}
To ensure geometric coherence, the generated dense projections $\hat{\mathcal{P}}_t$ at poses $\hat{\boldsymbol{\theta}}$ and the sparse measurements $\mathcal{P}$ at poses $\boldsymbol{\theta}$ must jointly correspond to a common 3D volume. Let $\mathcal{A}$ denote the forward Radon transform and $\mathcal{A}^\dagger$ its approximate inverse (FBP~\cite{shepp1974fourier} for parallel- and fan-beam, FDK~\cite{feldkamp1984practical} for cone-beam). We reconstruct an intermediate volume $\hat{\mathbf{y}}_{0|t}=\mathcal{A}^\dagger_{\tilde{\boldsymbol{\theta}}}(\tilde{\mathcal{P}}_t)$ from all projections $\tilde{\mathcal{P}}_t=\mathcal{P}\cup\hat{\mathcal{P}}_t$ at combined poses $\tilde{\boldsymbol{\theta}}=\boldsymbol{\theta}\cup\hat{\boldsymbol{\theta}}$. This intermediate volume can be decomposed into two components $\hat{\mathbf{y}}_{0|t}=\hat{\mathbf{y}}_{0|t}^C+\hat{\mathbf{y}}_{0|t}^R$, separately representing a reconstruction from consistent parts in $\tilde{\mathcal{P}}_t$ and a residual volume~\cite{anastasio2001comments}. To ensure $\hat{\mathbf{y}}_{0|t}^R\rightarrow\mathbf{0}$ as the denoising process progresses, our geometric coherency loss decomposes naturally into the following synergistic terms:
\begin{equation}
\label{eq:lgeo}
    \mathcal{L}_{\rm geo}=\underbrace{\|\mathcal{P}-\mathcal{A}_{\boldsymbol{\theta}}\hat{\mathbf{y}}_{0|t}\|^2_2}_{\text{measurement fidelity}}+\underbrace{\|\hat{\mathcal{P}}_t-\mathcal{A}_{\hat{\boldsymbol{\theta}}}\hat{\mathbf{y}}_{0|t}\|^2_2}_{\text{novel-view coherency}}+\mathcal{L}_{\rm TV}(\hat{\mathbf{y}}_{0|t}).
\end{equation}
The \emph{measurement fidelity} term penalizes the discrepancy between the true sparse measurements $\mathcal{P}$ and their reprojection from $\hat{\mathbf{y}}_{0|t}$ at the original angles $\boldsymbol{\theta}$, ensuring the volume implied by the generated views is faithful to the observations. The \emph{novel-view coherency} term enforces that the generated projections $\hat{\mathcal{P}}_t$ are themselves consistent with the same volume, pushing independently sampled denoising chains into alignment and preventing them from converging to mutually incompatible local anatomies. The total variation term $\mathcal{L}_{\rm TV}$ further regularizes local intensity variations in $\hat{\mathbf{y}}_{0|t}$~\cite{zang2021intratomo,cai2024radiative}. 
% Applying this loss during early denoising steps (when $\hat{\mathcal{P}}_t$ is noisy) is intentional: it shapes the denoising trajectory, steering independently initialized chains toward a globally coherent 3D interpretation rather than only correcting the final output.

\noindent\textbf{Physics-Informed Consistency Refinement}
$\mathcal{L}_{\rm geo}$ enforces consistency \emph{globally}: the generated views must reproject correctly from the reconstruction $\hat{\mathbf{y}}_{0|t}$. We pair this reconstruction-mediated constraint with a direct sinogram constraint derived from projection physics. In the continuous parallel-beam setting, valid Radon sinograms satisfy the Helgason--Ludwig Range Conditions~\cite{ludwig1966radon}. We use a finite-angle, finite-order version of these conditions as a differentiable surrogate, which penalizes moment residuals outside the range of a coherent object.
Consider the Radon transform operator $\mathcal{A}_{\hat{\boldsymbol\theta}}(r,\theta,z)$ on projection angles $\hat{\boldsymbol{\theta}}$ with $r,\theta,z$ representing a unit cylindrical coordinate system, its $n$-th moment curve $\mathcal{C}_n=\int_{-1}^1r^n\mathcal{A}_{\hat{\boldsymbol\theta}}(r,\theta,z)dr$ can be represented in a Fourier series:
\begin{equation}
    \mathcal{C}_n=\alpha_0(z)+\sum_{i=1}^\infty\alpha_{2i-1}(z)\cos i\theta+\alpha_{2i}(z)\sin i\theta.
\label{eq:cn}
\end{equation}
For fan-beam data, we apply this surrogate after standard rebinning to an approximate parallel-beam sinogram~\cite{dreike1976convolution,kachelriess2000advanced}. This surrogate enters only $\mathcal{L}_{\mathrm{phy}}$, which means $\mathcal{L}_{\mathrm{geo}}$ relies solely on the forward operator $A$ and its approximate inverse $A^\dagger$, and therefore applies unchanged to cone-beam and non-coplanar geometries.
The coefficient vector $\boldsymbol{\alpha}(z)$ for a slice at height $z$ whose $i$-th component is $\alpha_i(z)$ satisfies the following Range Conditions for any order $n$: the $i$-th component of $\boldsymbol{\alpha}(z)$ is zero whenever $\lceil\frac{i}{2}\rceil+n$ is odd or $i>2n$~\cite{ludwig1966radon,huang2017restoration}. Consequently, $\boldsymbol{\alpha}(z)$ is sparse and possesses only $n+1$ degrees of freedom. 

At each denoising step, we select a slice at height $z=Z$ and solve for this $(n+1)$-dimensional vector $\boldsymbol{\alpha}_0(Z)$ of non-zero coefficients in $\boldsymbol{\alpha}(z)$ via least squares. We construct a design matrix $\mathbf{A}_0\in\mathbb{R}^{N\times(n+1)}$ based on the parity of $n$: Denoting Vandermonde-like matrices of trigonometric basis $\mathbf{C}_{e},\mathbf{S}_{e},\mathbf{C}_{o},\mathbf{S}_{o}\in\mathbb{R}^{N\times\lceil\frac{n}{2}\rceil}$ whose $i$-th column vectors are $\cos 2i\hat{\boldsymbol{\theta}}, \sin 2i\hat{\boldsymbol{\theta}}, \cos (2i-1)\hat{\boldsymbol{\theta}}, \sin(2i-1)\hat{\boldsymbol{\theta}}$, $\mathbf{A}_0$ equals $[\mathbf{1}|\mathbf{C}_{e}|\mathbf{S}_{e}]$ when $n$ is even and $[\mathbf{C}_{o}|\mathbf{S}_{o}]$ when $n$ is odd. In practice, $(\mathbf{A}_0^\top\mathbf{A}_0)^{-1}$ is implemented with a numerically stable least-squares solver. The closed-form solution is:
\begin{equation}
\hspace{-7pt}
    \boldsymbol{\alpha}_0(Z)=(\mathbf{A}_0^\top\mathbf{A}_0)^{-1}\left(\mathbf{A}_0^\top\left(\int_{-1}^1r^n\hat{\mathcal{P}}_t(r,\theta,Z)dr\right)\right).
\label{alpha_z}
\hspace{-7pt}
\end{equation}
Our physics-informed consistency loss then refines the residual:
\begin{equation}
    \mathcal{L}_{\rm phy}=||\int_{-1}^1r^n\hat{\mathcal{P}_t}(r,\theta,Z)dr-\textbf{A}_0\boldsymbol{\alpha}_0(Z)||_2^2.
\end{equation}
This objective penalizes the distance from the generated views to the finite-order moment subspace. It complements $\mathcal{L}_{\rm geo}$ by discouraging sinogram patterns that fit the diffusion prior locally but violate global projection physics.

\noindent\textbf{Guided Denoising Update}
At each denoising step $t<T$, before performing reverse sampling with Eq.~\eqref{ddpminference}, we update each $\hat{\mathbf{x}}_t$ via gradient descent on the combined loss to steer its predicted clean projection $\hat{\mathbf{x}}_{0|t}$ toward geometric and physical consistency. The losses are evaluated with differentiable projection and reconstruction operators, so gradients flow through $\hat{\mathbf{x}}_{0|t}$ back to $\hat{\mathbf{x}}_t$. This guidance process is typically defined as~\cite{guo2024gradient}:
\begin{equation}
    \hat{\mathbf{x}}_{t}\leftarrow\hat{\mathbf{x}}_{t}-\gamma\cdot\nabla_{\hat{\mathbf{x}}_t}\mathcal{L},
\end{equation}
where $\gamma>0$ is a guidance scale (learning rate) and $\mathcal{L}$ is any loss function. In practice we use the Adam optimizer~\cite{kingma2014adam} with fixed learning rate $\gamma_0$ and moment decay rates $\beta_1,\beta_2$, which provides per-parameter adaptive gradient scaling for more stable convergence than vanilla gradient descent:
\begin{equation}
    \hat{\mathbf{x}}_{t}\leftarrow\texttt{Adam}\!\left(\hat{\mathbf{x}}_{t},\;\nabla_{\hat{\mathbf{x}}_t}(\lambda_1\mathcal{L}_{\rm geo}+\lambda_2\mathcal{L}_{\rm phy});\;\gamma_0,\beta_1,\beta_2\right),
\end{equation}
where $\lambda_1$, $\lambda_2$ are loss weights. This guided sampling process iteratively refines the ensemble of projections $\hat{\mathcal{P}}_t$, so that each final denoised output $\hat{\mathbf{x}}_0\in\hat{\mathcal{P}}_0$ follows the learned projection statistics and remains geometrically coherent with all projections. The final tomographic reconstruction $\hat{\mathbf{y}}_0$ is obtained by applying any standard reconstructor to the combined sinogram $\mathcal{P}\cup\hat{\mathcal{P}}_0$.

The guided denoising update resembles DPS~\cite{chung2022diffusion} and DDNM~\cite{wang2022zero} in its construction of the consistency objectives through the Tweedie's estimate $\hat{\mathbf{x}}_{0|t}$, and its Range-Null space decomposition of $\hat{\mathbf{y}}_{0|t}$ to get consistent and residual volumes. The key difference lies in the prior: DPS and DDNM assume a pretrained prior over the target (in our case, CT) domain, whereas ours is learned only from pose-conditioned projections and is paired with tomography-specific consistency objectives rather than generic data fidelity. We provide an algorithmic overview of this guided generation process in Appendix~\ref{sec:A}.

\section{Experiments}

\noindent\textbf{Dataset Curation} We evaluate on xSV-CT regime with two public datasets separately on the pulmonary (CT-RATE)~\cite{hamamci2024developing} and cranial (SinoCT)~\cite{hooper2021impact} cavities. We simulated 2D parallel-beam projections for CT-RATE using LEAP~\cite{kim2023differentiable} package. For SinoCT, we use the provided fan-beam sinograms, which were simulated with GE's CatSim simulator~\cite{catsim} to model a polychromatic source and detector response. These public datasets were built from dense acquisitions, which lets us compute metrics against dense references; no component of our method accesses more than one view per object during training. More details can be found in Appendix~\ref{sec:B}.

\noindent\textbf{Controlled xSV-CT Simulation} To evaluate against known ground truth while preserving the xSV-CT information pattern, we use these public CT datasets with three access rules: (i) at training, our model sees \textit{one} 2D projection per case together with its scanner pose; (ii) at inference, only $M\in\{1,4,8\}$ projections plus their poses are exposed; (iii) the dense 3D volume is held out from every method component of ours and used only as a reference for metric computation. It is important to note that, unlike our method, concurrent learning-based methods require reconstructed slice or volumes at training and thus do not fully respect the xSV-CT regime.

\noindent\textbf{Evaluation Metrics} 
Following prior works~\cite{lin2024c,barbano2025steerable}, we compute Peak Signal-to-Noise Ratio (PSNR) and Structural Similarity Index Measure (SSIM)~\cite{wang2004image} between ground truths and reconstructions. We also use Learned Perceptual Image Patch Similarity (LPIPS)~\cite{zhang2018unreasonable} to compare the perceptual authenticity of reconstructed volumes and projections. Reconstruction metrics are averaged over all axial slices and all images are rescaled to $[0,1]$ before metric computation.

\noindent\textbf{Implementation Details} We implement our method with PyTorch~\cite{paszke2019pytorch} and train it on a 4 NVIDIA RTX Ada 6000 GPU cluster. Our generative training uses a constant learning rate of $4\times 10^{-5}$ with a batch size of 8 per GPU and a projection size of $256\times256$ for CT-RATE and $64\times512$ for SinoCT. Training uses AdamW~\cite{loshchilov2017decoupled} with momentum 0.99 and weight decay $10^{-5}$. At inference, we select the degree of moment curve to be $n=20$ in Eq.~\eqref{alpha_z} with $\lambda_1=10^{-2}$ and $\lambda_2=1$. More details on dataset curation, method implementation and runtime analysis can be found in Appendix~\ref{sec:B}.

\subsection{Comparison Studies}
\begin{figure*}[h]
    \centering
    \vspace{-5pt}
    \includegraphics[width=\linewidth]{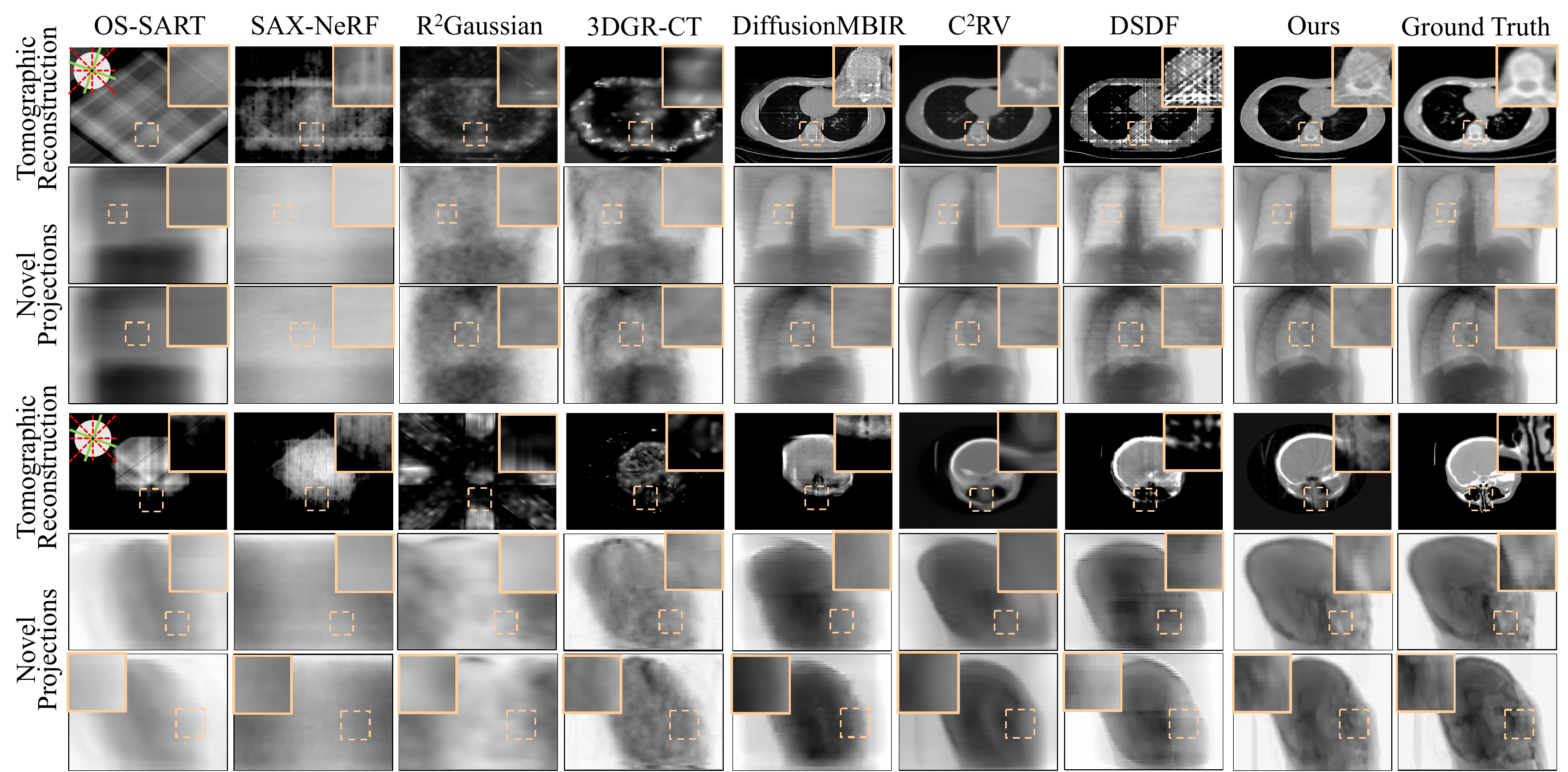}
    \caption{Qualitative comparison of tomographic reconstruction and novel view synthesis from extremely sparse (4) views on CT-RATE and SinoCT datasets. Input angles are marked in \textcolor{red}{red} dashed and novel angles in \textcolor[rgb]{0.57,0.81,0.31}{green} solid lines in the inset polar plots.
    }
    \label{fig:res}
    \vspace{-5pt}
\end{figure*}

\begin{table*}[t]
    \centering
    % \Huge
    \setlength{\tabcolsep}{1.mm}
    \caption{Quantitative comparison for novel view synthesis on a stratified subset of 50 test cases per dataset. Metrics are averaged across 180 synthesized novel projections covering the entire pose space. We highlight \colorbox{tabfirst}{best}, \colorbox{tabsecond}{second best} and \colorbox{tabthird}{third best} results.
    }
    \label{tab:compare1}
    \resizebox{\linewidth}{!}{\begin{tabular}{p{1cm}ll ccc ccc ccc ccc ccc ccc}
    \toprule
        \multicolumn{3}{c}{Metric} & \multicolumn{6}{c}{PSNR$\uparrow$} & \multicolumn{6}{c}{SSIM$\uparrow$} & \multicolumn{6}{c}{LPIPS$\downarrow$}  \\
        \cmidrule(lr){1-3}\cmidrule(lr){4-9}\cmidrule(lr){10-15}\cmidrule(lr){16-21}
        \multicolumn{3}{c}{Dataset}& \multicolumn{3}{c}{CT-RATE~\cite{hamamci2024developing}} & \multicolumn{3}{c}{SinoCT~\cite{hooper2021impact}}& \multicolumn{3}{c}{CT-RATE} & \multicolumn{3}{c}{SinoCT}& \multicolumn{3}{c}{CT-RATE} & \multicolumn{3}{c}{SinoCT}\\
        \cmidrule(lr){1-3}\cmidrule(lr){4-6}\cmidrule(lr){7-9}\cmidrule(lr){10-12}\cmidrule(lr){13-15}\cmidrule(lr){16-18}\cmidrule(lr){19-21}
        \multicolumn{3}{c}{Inference Views}&1&4&8&1&4&8&1&4&8&1&4&8&1&4&8&1&4&8\\
        \midrule
        \hspace{1.2em}\multirow{5}{*}{\rotatebox[origin=c]{90}{Projection-}}&{\hspace{-1.2em}\multirow{5}{*}{\rotatebox[origin=c]{90}{based}}}&FBP~\cite{shepp1974fourier} &7.2&10.1&11.3&10.6&11.5 &12.2&0.15&0.13&0.29&0.28&0.39&0.42&0.50&0.48&0.48&0.52&0.48&0.43\\
        % CGLS~\cite{scales1987tomographic} & &&&16.3\\
        &&OS-SART~\cite{wang2004ordered} &7.4&8.9&14.7&12.1&11.9&19.6&0.14&0.16&0.31&0.35&0.44&0.53&0.50&0.46&0.36&0.54&0.43&0.40\\
        \cmidrule(r){3-21}
        &&SAX-NeRF~\cite{cai2024structure} &15.0&19.7&20.5&11.4&12.3&15.1&0.32&0.56&0.61&0.28&0.42&0.49&{0.43}&0.42&0.39&0.52&0.43&0.40\\
        &&R$^2$-Gaussian~\cite{zha2024r} &14.1&15.4&25.8&11.2&{13.6}&16.8&0.46&0.52&0.59&0.21&0.31&0.44&0.45&0.45&0.40&0.58&0.51&0.46\\
        &&3DGR-CT~\cite{li20253dgr}& 15.0&\cellcolor{tabthird}{24.6}&26.8&11.5&12.6&16.1&0.48&0.51&0.61&0.28&{0.44}&0.49&0.45&0.40&0.37&0.49&0.46&\cellcolor{tabthird}{0.39}\\
        \cmidrule(lr){1-21}
        \hspace{1.2em}\multirow{3}{*}{\rotatebox[origin=c]{90}{Volume-}}&\hspace{-1.2em}\multirow{3}{*}{\rotatebox[origin=c]{90}{based}}&DiffusionMBIR~\cite{chung2023solving}&{17.5}&23.1&\cellcolor{tabthird}{29.8}&{13.0}&13.3&\cellcolor{tabthird}{20.1}&0.50&{0.76}&{0.84}&0.35&0.37&{0.56}&0.52&{0.35}&\cellcolor{tabsecond}{0.21}&\cellcolor{tabsecond}{0.45}&\cellcolor{tabsecond}{0.40}&\cellcolor{tabsecond}{0.38}\\
        &&C$^2$RV~\cite{lin2024c}&\cellcolor{tabthird}{20.5}&\cellcolor{tabsecond}{27.8}&\cellcolor{tabsecond}{33.5}&\cellcolor{tabthird}{17.4}&\cellcolor{tabthird}{17.8}&22.8&\cellcolor{tabthird}{0.74}&\cellcolor{tabsecond}{0.89}&\cellcolor{tabsecond}{0.94}&\cellcolor{tabthird}{0.51}&\cellcolor{tabthird}{0.58}&\cellcolor{tabthird}{0.67}&\cellcolor{tabfirst}{0.32}&\cellcolor{tabsecond}{0.25}&{0.24}&0.49&\cellcolor{tabthird}{0.42}&0.39\\
        &&DSDF~\cite{pan2025draw}&\cellcolor{tabsecond}{21.1}&{23.6}&{27.3}&\cellcolor{tabsecond}{18.1}&\cellcolor{tabsecond}{18.1}&\cellcolor{tabsecond}{20.6}&\cellcolor{tabsecond}{0.75}&\cellcolor{tabthird}{0.81}&\cellcolor{tabthird}{0.90}&\cellcolor{tabsecond}{0.56}&\cellcolor{tabsecond}{0.59}&\cellcolor{tabsecond}{0.69}&\cellcolor{tabthird}{0.39}&\cellcolor{tabthird}{0.32}&\cellcolor{tabthird}{0.21}&\cellcolor{tabthird}{0.48}&0.45&0.40\\
        \cmidrule(lr){1-21}
        &&Ours&\cellcolor{tabfirst}{23.1}&\cellcolor{tabfirst}{32.8}&\cellcolor{tabfirst}{36.3}&\cellcolor{tabfirst}{19.6}&\cellcolor{tabfirst}{24.3}&\cellcolor{tabfirst}{25.7}&\cellcolor{tabfirst}{0.80}&\cellcolor{tabfirst}{0.93}&\cellcolor{tabfirst}{0.95}&\cellcolor{tabfirst}{0.63}&\cellcolor{tabfirst}{0.74}&\cellcolor{tabfirst}{0.77}&\cellcolor{tabsecond}{0.34}&\cellcolor{tabfirst}{0.23}&\cellcolor{tabfirst}{0.17}&\cellcolor{tabfirst}{0.43}&\cellcolor{tabfirst}{0.38}&\cellcolor{tabfirst}{0.34}\\
    \bottomrule
        
    \end{tabular}}
\vspace{-5pt}
\end{table*}

\begin{table*}[t]
    \centering
    % \Huge
    \setlength{\tabcolsep}{1.mm}
    \caption{Quantitative comparison for tomographic reconstruction on the same 50-case stratified subset as Table~\ref{tab:compare1}.
    }
    \label{tab:compare2}
    \resizebox{\linewidth}{!}{    \begin{tabular}{p{1cm}ll ccc ccc ccc ccc ccc ccc}
    \toprule
        \multicolumn{3}{c}{Metric} & \multicolumn{6}{c}{PSNR$\uparrow$} & \multicolumn{6}{c}{SSIM$\uparrow$} & \multicolumn{6}{c}{LPIPS$\downarrow$}  \\
        \cmidrule(lr){1-3}\cmidrule(lr){4-9}\cmidrule(lr){10-15}\cmidrule(lr){16-21}
        \multicolumn{3}{c}{Dataset}& \multicolumn{3}{c}{CT-RATE~\cite{hamamci2024developing}} & \multicolumn{3}{c}{SinoCT~\cite{hooper2021impact}}& \multicolumn{3}{c}{CT-RATE} & \multicolumn{3}{c}{SinoCT}& \multicolumn{3}{c}{CT-RATE} & \multicolumn{3}{c}{SinoCT}\\
        \cmidrule(lr){1-3}\cmidrule(lr){4-6}\cmidrule(lr){7-9}\cmidrule(lr){10-12}\cmidrule(lr){13-15}\cmidrule(lr){16-18}\cmidrule(lr){19-21}
        \multicolumn{3}{c}{Inference Views}&1&4&8&1&4&8&1&4&8&1&4&8&1&4&8&1&4&8\\
        \midrule
        \hspace{1.2em}\multirow{5}{*}{\rotatebox[origin=c]{90}{Projection-}}&{\hspace{-1.2em}\multirow{5}{*}{\rotatebox[origin=c]{90}{based}}}&FBP~\cite{shepp1974fourier}&7.4&8.0&9.6&11.7&11.7&13.0&0.12&0.19&0.23&0.34&0.38&0.39&0.55&0.53&0.54&0.54&0.45&0.43 \\
        &&OS-SART~\cite{wang2004ordered} &7.9&7.8&11.2&12.7&17.8&19.0&0.15&0.19&0.27&0.33&0.37&0.44&0.50&0.48&0.56&0.53&0.43&0.41\\
        \cmidrule(r){3-21}
        &&SAX-NeRF~\cite{cai2024structure}&6.3&11.4&20.2&9.4&15.0&17.9&0.15&0.37&0.54&0.29&0.40&0.46&0.66&0.52&0.40&0.49&0.42&0.36 \\
        &&R$^2$-Gaussian~\cite{zha2024r}&9.8&15.9&24.8&9.3&12.0&19.1& 0.21&0.40&0.56&0.23&0.28&0.42&0.55&0.45&{0.29}&0.56&0.47&0.41\\
        &&3DGR-CT~\cite{li20253dgr} &\cellcolor{tabthird}{11.7}&19.6&26.2&9.1&15.4&20.3&0.21&0.38&0.53&0.33&0.34&0.53&0.51&0.41&0.31&\cellcolor{tabthird}{0.48}&0.45&0.40\\
        \cmidrule(lr){1-21}
        
        \hspace{1.2em}\multirow{3}{*}{\rotatebox[origin=c]{90}{Volume-}}&\hspace{-1.2em}\multirow{3}{*}{\rotatebox[origin=c]{90}{based}}&DiffusionMBIR~\cite{chung2023solving}&\cellcolor{tabsecond}{16.4}&21.8&\cellcolor{tabsecond}{27.1}&\cellcolor{tabsecond}{16.3}&\cellcolor{tabfirst}{21.5}&\cellcolor{tabthird}{24.9}&0.20&\cellcolor{tabthird}{0.44}&{0.60}&\cellcolor{tabthird}{0.59}&\cellcolor{tabfirst}{0.72}&\cellcolor{tabsecond}{0.80}&\cellcolor{tabthird}{0.47}&0.40&\cellcolor{tabsecond}{0.27}&0.48&\cellcolor{tabthird}{0.41}&\cellcolor{tabsecond}{0.34}\\
        &&C$^2$RV~\cite{lin2024c}&11.0&\cellcolor{tabsecond}{23.3}&26.6&\cellcolor{tabthird}{13.3}&\cellcolor{tabthird}{19.6}&\cellcolor{tabfirst}{25.9}&\cellcolor{tabsecond}{0.28}&\cellcolor{tabsecond}{0.50}&\cellcolor{tabthird}{0.64}&\cellcolor{tabsecond}{0.63}&\cellcolor{tabthird}{0.71}&0.76&0.54&\cellcolor{tabthird}{0.39}&\cellcolor{tabfirst}{0.25}&0.50&0.45&0.38\\
        &&DSDF~\cite{pan2025draw}&9.3&\cellcolor{tabthird}{22.1}&\cellcolor{tabthird}{27.0}&10.4&17.8&23.9&\cellcolor{tabthird}{0.23}&{0.43}&\cellcolor{tabsecond}{0.64}&0.40&\cellcolor{tabsecond}{0.71}&\cellcolor{tabthird}{0.77}&\cellcolor{tabsecond}{0.43}&\cellcolor{tabsecond}{0.36}&0.30&\cellcolor{tabsecond}{0.48}&\cellcolor{tabsecond}{0.36}&\cellcolor{tabthird}{0.35}\\

        &&Ours&\cellcolor{tabfirst}{21.1}&\cellcolor{tabfirst}{24.5}&\cellcolor{tabfirst}{28.6}&\cellcolor{tabfirst}{16.4}&\cellcolor{tabsecond}{20.0}&\cellcolor{tabsecond}{25.2}&\cellcolor{tabfirst}{0.46}&\cellcolor{tabfirst}{0.51}&\cellcolor{tabfirst}{0.69}&\cellcolor{tabfirst}{0.66}&0.70&\cellcolor{tabfirst}{0.80}&\cellcolor{tabfirst}{0.41}&\cellcolor{tabfirst}{0.32}&\cellcolor{tabthird}{0.27}&\cellcolor{tabfirst}{0.47}&\cellcolor{tabfirst}{0.36}&\cellcolor{tabfirst}{0.34}\\
    \bottomrule
        
    \end{tabular}}
\vspace{-5pt}
\end{table*}

% \begin{figure}
%     \centering
%     \includegraphics[width=0.6\linewidth]{figs/result2.pdf}
%     \caption{Tomographic reconstruction across different xSV-CT settings comparing our method against two best-performing methods~\cite{pan2025draw,chung2023solving}. One axial and one coronal slice is shown for each case.
%     Our method produces more plausible reconstructions with fewer artifacts across all sparsity levels.
%     }
%     \label{fig:res2}
% % \vspace{-15pt}
% \end{figure}

We compare against 5 optimization-based methods, including classical analytical (FBP~\cite{shepp1974fourier}), classical iterative (OS-SART~\cite{wang2004ordered}), and NeRF/Gaussian-volume methods (SAX-NeRF~\cite{cai2024structure}, R$^2$-Gaussian~\cite{zha2024r}, 3DGR-CT~\cite{li20253dgr}), as well as 3 learning-based methods, including a diffusion method trained on axial slices (DiffusionMBIR~\cite{chung2023solving}) and two regression methods trained on dense 3D volumes (C$^2$RV~\cite{lin2024c}, DSDF~\cite{pan2025draw}).
We evaluate the performance of each method on two tasks, \textbf{1}) 2D novel view synthesis and \textbf{2}) 3D tomographic reconstruction, with an extremely sparse set of $M \in \{1, 4, 8\}$ input views. To ensure fairness, we retrain each method according to their per-dataset or per-viewcount training schemes. More details can be found in Appendix~\ref{sec:B}.

\noindent\textbf{Novel View Synthesis} As shown in Table~\ref{tab:compare1}, analytical reconstruction is strongly affected by the missing angular coverage in xSV-CT. Other optimization-based methods improve over FBP and OS-SART but degrade as $M$ decreases. Learning-based methods maintain reasonable quality at this sparsity level using priors trained from dense-view reconstructions. Our method achieves the best projection metrics while using one projection per training sample.

\noindent\textbf{Tomographic Reconstruction} As shown in Table~\ref{tab:compare2}, the performance hierarchy observed for novel view synthesis (Table~\ref{tab:compare1}) generally persists: learning-based methods outperform optimization-based approaches, and our method achieves comparable or superior results to learning-based methods that require densely scanned training data. Qualitative results are shown in Fig.~\ref{fig:res} (4-view), with more inference-view counts and stress-test analysis in Appendix~\ref{sec:E}. The gains are concentrated in the xSV-CT regime ($M{\leq}8$) and narrow as $M$ grows into the standard sparse-view setting (1--64 input views; see Appendix~\ref{sec:C}). This behavior is desirable: the learned prior is most useful when measurements are scarce, and becomes less central when angular coverage itself carries enough information. When more projections per sample are available at training time, performance improves smoothly (Appendix~\ref{sec:C}); single-projection training is the most data-efficient end of this trade-off. 

\begin{figure*}[t]
    \centering
    \includegraphics[width=\linewidth]{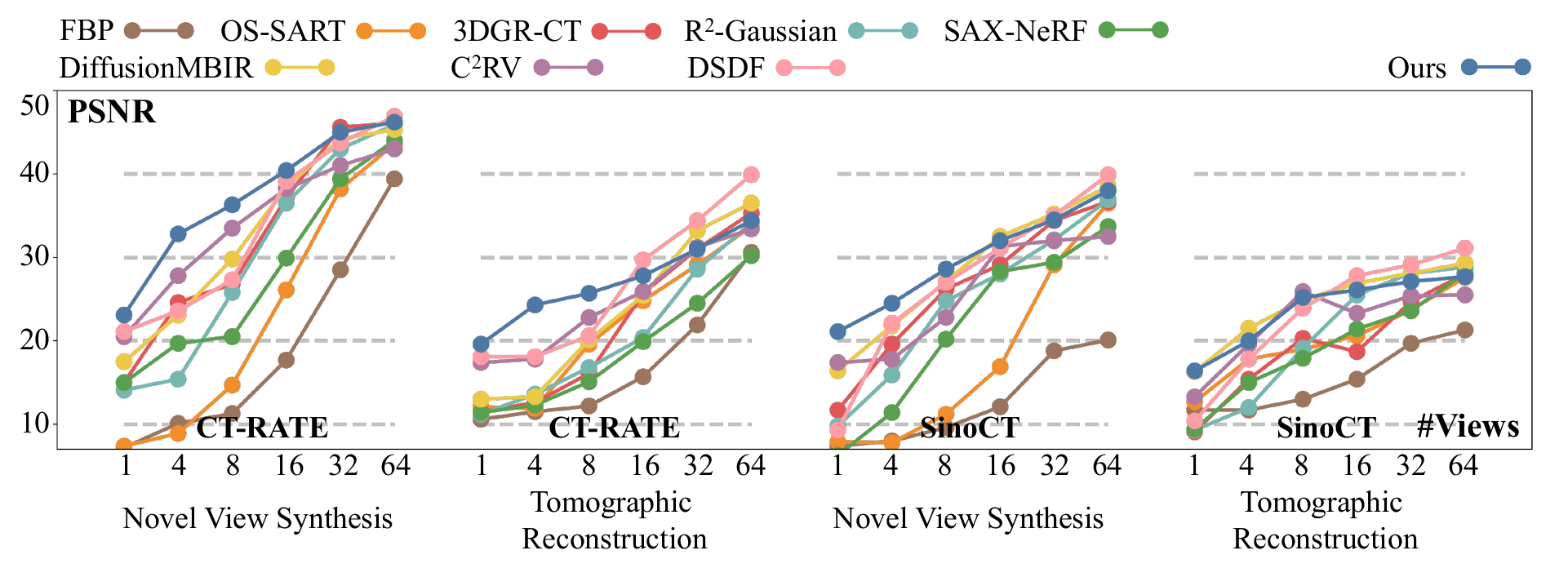}
    \caption{Performance scaling with the number of input views, metrics averaged over 30 cases.}
    \label{fig:moreviews}
    \vspace{-15pt}
\end{figure*}
\noindent\textbf{Inference Performance Scaling with Number of Input Views}
We evaluate how performance scales as the number of input views $M$ grows from 1 to 64, which covers both the xSV-CT regime ($M{\leq}8$) and the standard sparse-view setting. As shown in Fig.~\ref{fig:moreviews}, our method performs best for $M{\leq}8$ and remains competitive with learning-based methods as $M$ increases, with the gap narrowing as more views become available. This contrasts with earlier findings that projection-domain diffusion priors underperform slice-domain priors because errors in independently generated projections accumulate during reconstruction~\cite{su2026diffnr}. Our ablation reproduces that failure when both guidance losses are removed (Table~\ref{tab:ab1}); coupling all generated views to a shared volume and to the Radon range during sampling largely prevents it, making projection-domain priors a preferable choice when dense acquisition itself is the constraint.

% \noindent\textbf{Runtime Analysis} End-to-end runtime scales with the number of synthesized views $N$ and DDIM steps. At 4-view input with $N{=}180$ novel projections and 100 DDIM steps, inference takes ${\sim}46$ min on one RTX Ada 6000 GPU. Reducing to $N{=}90$ provides ${\sim}2{\times}$ speedup with only ${\sim}12.6\%$ PSNR degradation (``fast mode''). Computation is dominated by diffusion sampling; the guidance losses (Sec.~\ref{sec:3.2}) add negligible overhead ($<$5\% of total runtime). This compares favourably to DiffusionMBIR, which exceeds 10 hours per test case at equivalent quality.

% \noindent\textbf{Hallucination Analysis} Under extreme sparsity, learned priors inevitably generate structure not fully constrained by measurements. Two mechanisms in our method bound this risk: the measurement fidelity term of $\mathcal{L}_{\rm geo}$ (Eq.~\eqref{eq:lgeo}) directly penalises deviations at measured angles, so the diffusion chain is anchored to the actual observations; and $\mathcal{L}_{\rm phy}$ enforces global range-condition consistency, preventing hallucinated structures from persisting across the full sinogram. We characterise failure modes qualitatively in the Supplementary Material, covering noisy-input, angular-uncertainty, and out-of-distribution settings. Under these conditions, our method still produces reconstructions closer to the ground truth than the best-performing baselines.

\begin{table}[t]
    \centering\scriptsize
    \setlength{\tabcolsep}{1pt}
    \caption{Ablation study on key components evaluated with 8 input views on 50 test cases per dataset. }
    \label{tab:ab1}
    \resizebox{0.8\linewidth}{!}{
    % \Huge
    {
    \setlength{\heavyrulewidth}{0.11em}
    \begin{tabular}{l cccccc cccccc}
        \toprule
        \multicolumn{1}{c}{Task} & \multicolumn{6}{c}{Novel View Synthesis} & \multicolumn{6}{c}{Tomographic Reconstruction} \\
        \cmidrule(lr){1-1}\cmidrule(lr){2-7}\cmidrule(lr){8-13}
        \multicolumn{1}{c}{Dataset}&\multicolumn{3}{c}{CT-RATE}&\multicolumn{3}{c}{SinoCT}&\multicolumn{3}{c}{CT-RATE}&\multicolumn{3}{c}{SinoCT}\\
         \cmidrule(lr){1-1}\cmidrule(lr){2-4}\cmidrule(lr){5-7}\cmidrule(lr){8-10}\cmidrule(lr){11-13}
        \multicolumn{1}{c}{Metric}&PSNR&SSIM&LPIPS&PSNR&SSIM&LPIPS&PSNR&SSIM&LPIPS&PSNR&SSIM&LPIPS\\
        \midrule
        No pose&11.9&0.21&0.56&12.5&0.23&0.50&9.4&0.17&0.61&10.7&0.25&0.53\\
        Naive CrossAttn&24.5&0.40&0.48&20.2&0.45&0.47&11.5&0.33&\cellcolor{tabthird}{0.41}&15.6&0.39&0.50\\
        \cmidrule(lr){1-13}
        w.o. $\mathcal{L}_{\rm phy}$, $\mathcal{L}_{\rm geo}$&15.6&0.32&0.55&14.6&0.32&0.49&9.2&0.19&0.59&10.0&0.18&0.53\\
        w.o. $\mathcal{L}_{\rm geo}$&\cellcolor{tabthird}{28.6}&\cellcolor{tabthird}{0.78}&\cellcolor{tabthird}{0.30}&\cellcolor{tabthird}{21.4}&\cellcolor{tabthird}{0.50}&\cellcolor{tabthird}{0.40}&\cellcolor{tabthird}{16.4}&\cellcolor{tabthird}{0.35}&0.44&\cellcolor{tabthird}{18.9}&\cellcolor{tabthird}{0.51}&\cellcolor{tabsecond}{0.47}\\
        w.o. $\mathcal{L}_{\rm phy}$&\cellcolor{tabsecond}{30.1}&\cellcolor{tabsecond}{0.88}&\cellcolor{tabsecond}{0.26}&\cellcolor{tabsecond}{22.1}&\cellcolor{tabsecond}{0.56}&\cellcolor{tabsecond}{0.39}&\cellcolor{tabsecond}{22.0}&\cellcolor{tabsecond}{0.44}&\cellcolor{tabsecond}{0.39}&\cellcolor{tabsecond}{19.1}&\cellcolor{tabsecond}{0.56}&\cellcolor{tabthird}{0.47}\\
        \cmidrule(lr){1-13}
        Ours (full)&\cellcolor{tabfirst}{36.3}&\cellcolor{tabfirst}{0.95}&\cellcolor{tabfirst}{0.17}&\cellcolor{tabfirst}{25.7}&\cellcolor{tabfirst}{0.77}&\cellcolor{tabfirst}{0.34}&\cellcolor{tabfirst}{28.6}&\cellcolor{tabfirst}{0.69}&\cellcolor{tabfirst}{0.27}&\cellcolor{tabfirst}{25.2}&\cellcolor{tabfirst}{0.80}&\cellcolor{tabfirst}{0.34}\\
        \bottomrule
    \end{tabular}
}
    }
\vspace{-10pt}
\end{table}

\begin{figure}[t]
    \centering
    \includegraphics[width=0.8\linewidth,clip,trim={0 0 0 0}]{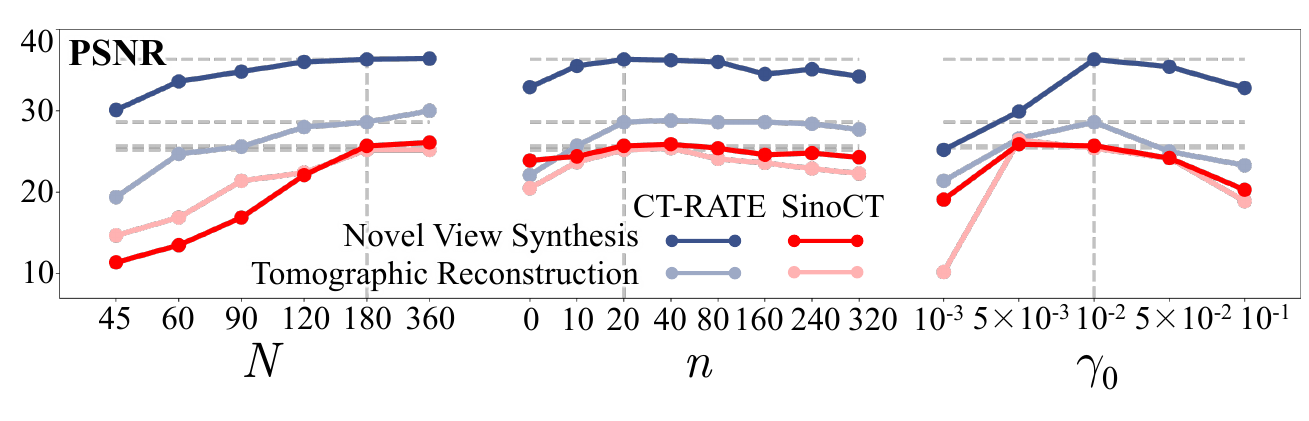}
    \vspace{-10pt}
    \caption{Parameter sensitivity analysis with 8 input views. Blue and red curves show PSNR on CT-RATE and SinoCT datasets respectively. Gray dashed lines indicate our default settings.
    }
    \label{fig:ab1}
\vspace{-15pt}
\end{figure}

\subsection{Ablation Studies}
Table~\ref{tab:ab1} ablates our key design choices. 
First, we evaluate our pose-aware projection-domain diffusion prior by comparing against (\textbf{1}) removing pose information entirely, and (\textbf{2}) using direct cross-attention on raw pose angles and the diffusion latents. Without pose information, the model generates uncalibrated projections even when guided by the consistency losses. Raw-angle cross-attention also underperforms because it lacks the periodic and reflection-related structure of projection angles. Second, we evaluate consistency losses, $\mathcal{L}_{\rm geo}$ and $\mathcal{L}_{\rm phy}$, during inference. With the pose-aware projection-domain prior fixed, either loss improves reconstruction, and using both losses gives the best results across both datasets. This supports the intended division of labor: the diffusion model proposes pose-conditioned projections, while the guidance losses align them with measured views and projection physics.

% \begin{figure}[t]
%     \centering
%     \includegraphics[width=0.7\linewidth]{figs/robust.pdf}
%     \caption{Qualitative evaluation under noisy input projections. Each row shows reconstruction from 8 noisy projections (one example shown in the first row) under simulated 75\%, 50\%, 25\% and 10\% of original X-ray dosage to guide the inference process.}
%     \label{fig:robust}
% \vspace{-10pt}
% \end{figure}

\subsection{Robustness Analysis}
We present the robustness of our model against key hyperparameters and faithfulness to sparse measurements in this section, with more analysis and failure cases provided in Appendices~\ref{sec:C} and~\ref{sec:D}. 
\noindent\textbf{Parameter Sensitivity} Fig.~\ref{fig:ab1} analyzes three key hyperparameters: number of novel projections $N$ generated at inference, the order $n$ of the moment curve used in Eq.~\eqref{alpha_z}, and the guidance scale $\gamma_0$ for inference losses. Generating more novel projections generally improves quality at the cost of longer inference, as a denser view set provides more stable solutions for Eq.~\eqref{alpha_z} and improves the reconstruction quality of the FBP operator $\mathcal{A}^\dagger$. We select $N=180$ as our default to balance reconstruction quality and computational cost. 
The moment order $n$ works well across a wide range (20$\sim$80), degrading only when $n$ is too small to describe the sinogram moment structure or too large for stable optimization. 
The optimal $\gamma_0$ differs between datasets, likely because SinoCT uses a higher axial resolution, which introduces more fine-grained variations across projections than CT-RATE, thus requiring a smaller guidance scale for refining the generated projections.

\label{sec:faithfulness}
\noindent\textbf{Measurement Anchoring Analysis}
Under 1- and 4-view inputs, generated angles whose nearest measured view is more than ${\sim}30^\circ$ away draw more strongly on the learned prior. We quantify this measurement dependence with two checks. \begin{wraptable}{r}{0.35\textwidth}
  \centering
  \vspace{-5pt}
  \setlength{\tabcolsep}{1mm}
  \resizebox{\linewidth}{!}{
  \begin{tabular}{lcc}
    \toprule
    Views & \textit{mean}$\times10^{-3}$ & \textit{std.}$\times10^{-4}$ \\
    \midrule
    1-view   &  5.48& 4.64\\
    4-view   &  1.31 & 1.74\\
    8-view   &   0.96 &0.91\\
    \bottomrule
  \end{tabular}}
  \caption{Cross-seed $\ell_2$ residual on generated views.}
  \label{tab:wrapped}\vspace{-15pt}
\end{wraptable}\emph{(i) Cross-seed repeatability.} Running our method with multiple random seeds yields similar projections across runs: mean cross-seed $\ell_2$ residuals are $5.48{\times}10^{-3}$,
$1.31{\times}10^{-3}$, and $0.96{\times}10^{-3}$ at 1-, 4-, and 8-view inputs, with standard deviations around $10^{-4}$ (Table~\ref{tab:wrapped}). \emph{(ii) Angular-distance dependence.} Residuals on generated projections grow with angular distance from the nearest input view (see Appendix~\ref{sec:C}), matching the expected geometry: views closer to measurements are more directly constrained.

% \noindent\textbf{Noisy Projection} Following the ``low-dose'' simulation strategy of noisy X-ray images in~\cite{yu2012development,vzabic2013low}, we use TIGRE~\cite{biguri2016tigre} to inject Poisson and Gaussian noises into the given views to simulate the effect of photon scattering and electronic static that affect the quality of projections under X-ray outputs with lower dosages. As shown in Table~\ref{tab:robust} (left), thanks to our strong diffusion prior, noisy input projections do not significantly influence the overall structures with graceful degradation of fine anatomical details. 
% % This demonstrates the robustness of learning from clean 2D projections during training.

% \noindent\textbf{Angular Uncertainty} We introduce pose uncertainty by randomly perturbing each input angle $\theta\in\boldsymbol{\theta}$ within a certain range $\Delta\theta$. Table~\ref{tab:robust} (right) shows that our approach outperforms comparisons with angular errors within $\pm2.5^\circ$ and remains competitive up to $\pm7.5^\circ$, showing tolerance to imprecise angle calibration. We also present visualizations for robustness of different methods in the Appendix.

% Ablations:
% Tab.:
% 1. no pose; 
% 2. use cross-attn for pose; 
% 3. no reconstruction model at generation;
% 4. no consistency loss at generation;

% Fig.:
% Effect of different number of view

\section{Conclusion}
We introduced a train-sparse / infer-sparse framework for xSV-CT, where acquisition limits constrain both reconstruction and training-data curation. We show that a projection-domain diffusion prior learned from single-view data can become useful when it is embedded inside a physics-guided sampling loop. Through this projection-domain prior, shared-volume reprojection, and finite-order Range Condition guidance, the method synthesizes tomographically coherent projections for standard reconstruction without dense-volume supervision in the target regime. Controlled experiments on two public CT datasets support this claim across 1-, 4-, and 8-view settings. We analyze the broader impacts of our method as well as the xSV-CT regime in Appendix~\ref{sec:F}. 
\textbf{Limitations:} We use public dense CT data to provide paired references for controlled evaluation; prospective sparse-only clinical or specimen studies are left for future work. Generated angles far from measured views draw more strongly on the learned prior, and our finite-order Range-Condition surrogate currently covers parallel-beam or rebinned fan-beam data. These points motivate calibrated trust maps and native non-parallel constraints such as Grangeat's Condition~\cite{grangeat2006mathematical}.

\section*{Acknowledgements}
The research is made possible under the JST-Mirai program from Japanese Science and Technology Agency with grant number JPMJMI24G1. The authors declare no competing interests.

\clearpage
\bibliographystyle{unsrt}
\bibliography{main}

\appendix
% \clearpage
% \setcounter{page}{1}
% \maketitle

% \renewcommand{\thesection}{\Alph{section}}
% \renewcommand{\thetable}{\Alph{table}}
% \renewcommand{\thefigure}{\Alph{figure}}
% \renewcommand{\theequation}{\Alph{equation}}
% \setcounter{section}{0}
% \setcounter{table}{4}

% \textcolor{red}{Title?}
\clearpage

This supplementary material provides: an algorithmic overview of our guided generation procedure, additional experimental details, extended analysis, failure analysis, potential societal impacts, and comprehensive qualitative results.
We also provide an accompanying \textbf{video} that visualizes 3D reconstructions across compared methods and ours for all cases presented in this appendix. The code for training and evaluation as well as model checkpoints will be available in: https://github.com/OvO1111/xSV-CT.

\section{Algorithmic Overview}
\label{sec:A}
We provide an overview of our \textit{Consistency-aware Guided Generation} process in Algorithm~\ref{alg:guided_generation}. Intuition and derivation of each component are described in Sec.~\ref{sec:3.2} of the main text.

\begin{algorithm}[h!]
\caption{Consistency-Aware Guided Generation}
\label{alg:guided_generation}
    \begin{algorithmic}[1]
    
    \Require Sparse projections $\mathcal{P}$ at poses $\boldsymbol{\theta}$; novel poses $\hat{\boldsymbol{\theta}}$; trained diffusion model $p_\phi$; Radon operator $\mathcal{A}$ and approximate inverse (via FBP~\cite{shepp1974fourier} or FDK~\cite{feldkamp1984practical}) $\mathcal{A}^{\dagger}$; loss weights $\lambda_1,\lambda_2$; Adam parameters $(\gamma_0,\beta_1,\beta_2)$; Diffusion schedules ($\bar{\alpha}_t$, $\alpha_t$, $\beta_t$)
    \Ensure Dense projections $\hat{\mathcal{P}}_0$ and reconstructed volume $\hat{\mathbf{y}}_0$.
    
    \State Initialize noisy projections $\hat{\mathcal{P}}_T = \{\hat{\mathbf{x}}_T^i \sim \mathcal{N}(0,I)\}_{i=1}^{N}$
    
    \For{$t = T, T-1, \dots, 1$}
    
        \For{each pose $\hat{\theta}_i \in \hat{\boldsymbol{\theta}}$}
    
            \State Predict clean projection
            \(
            \hat{\mathbf{x}}_{0|t}^i =
            (\hat{\mathbf{x}}_t^i - \sqrt{1-\bar{\alpha}_t}\,
            p_\phi(\hat{\mathbf{x}}_t^i;\hat{\theta}_i,t)) / \sqrt{\bar{\alpha}_t}
            \)
    
        \EndFor
    
        \State Collect $\hat{\mathcal{P}}_t=\{\hat{\mathbf{x}}_{0|t}^i\}_{i=1}^N$, $\tilde{\mathcal{P}}_t=\mathcal{P} \cup \hat{\mathcal{P}}_t$, $\tilde{\boldsymbol{\theta}}=\boldsymbol{\theta}\cup\hat{\boldsymbol{\theta}}$
    
        \State Reconstruct intermediate volume
        \(
        \hat{\mathbf{y}}_{0|t} = \mathcal{A}_{\tilde{\boldsymbol{\theta}}}^{\dagger}(\tilde{\mathcal{P}}_t)
        \)
    
        \State Compute 
        \(
        \mathcal{L}_{\text{geo}} =
        \|\mathcal{P}-\mathcal{A}_{\boldsymbol{\theta}}\hat{\mathbf{y}}_{0|t}\|_2^2
        +\|\hat{\mathcal{P}}_t-\mathcal{A}_{\hat{\boldsymbol{\theta}}}\hat{\mathbf{y}}_{0|t}\|_2^2
        +\mathcal{L}_{TV}(\hat{\mathbf{y}}_{0|t})
        \)
    
        \State Construct design matrix $\mathbf{A}_0$ at height $z=Z$, rebin $\hat{\mathcal{P}}_t$ for fan-beam geometries and solve Range Condition coefficients
        \(
        \boldsymbol{\alpha}_0(Z) =
        (\mathbf{A}_0^\top\mathbf{A}_0)^{-1}\mathbf{A}_0^\top\int_{-1}^{1} r^n \hat{\mathcal{P}}_t(r,\theta,Z) dr
        \)
    
        \State Compute 
        \(
        \mathcal{L}_{\text{phy}} = ||\int_{-1}^1r^n\hat{\mathcal{P}_t}(r,\theta,Z)dr-\textbf{A}_0\boldsymbol{\alpha}_0(Z)||_2^2
        \)
    
        \For{each projection $\hat{\mathbf{x}}_t^i$}
    
            \State Guided update
            \(
            \hat{\mathbf{x}}_t^i \leftarrow
            \texttt{Adam}\!\left(
            \hat{\mathbf{x}}_t^i,\,
            \nabla_{\hat{\mathbf{x}}_t^i}\left(\lambda_1 \mathcal{L}_{\text{geo}} + \lambda_2 \mathcal{L}_{\text{phy}}\right);
            \gamma_0,\beta_1,\beta_2
            \right)
            \)
    
            \State Reverse diffusion step
            \(
            \hat{\mathbf{x}}_{t-1}^i =
            \frac{1}{\sqrt{\alpha_t}}
            \left(
            \hat{\mathbf{x}}_t^i -
            \frac{\beta_t}{\sqrt{1-\bar{\alpha}_t}}
            p_\phi(\hat{\mathbf{x}}_t^i;\hat{\theta}_i,t)
            \right)
            + \beta_t \boldsymbol{\epsilon}
            \)
    
        \EndFor
    
    \EndFor
    
    \State Generated consistent projections $\hat{\mathcal{P}}_0 = \{\hat{\mathbf{x}}_0^i\}_{i=1}^N$
    
    \State Final reconstruction
    \(
    \hat{\mathbf{y}}_0 = \mathcal{A}_{\tilde{\boldsymbol{\theta}}}^{\dagger}(\mathcal{P} \cup \hat{\mathcal{P}}_0)
    \)
    
    \end{algorithmic}
\end{algorithm}

\section{Additional Experimental Details}
\label{sec:B}
\noindent\textbf{Dataset Configuration}
We list the projection parameters and dataset statistics in Table~\ref{tab:ds}. To construct the training and evaluation sets, we generated 3D volume projections from the CT-RATE~\cite{hamamci2024developing} dataset using a parallel-beam geometry. For the SinoCT~\cite{hooper2021impact} dataset, we utilized the existing fan-beam projections simulated with CatSim~\cite{catsim} (source-isocenter distance of 541\,mm, a source-detector distance of 949\,mm) provided by the original paper and resized them to a consistent shape. We selected 6{,}000 cases from CT-RATE and 4{,}000 from SinoCT based on the number of slices and image intensity range, discarding CTs from the same patient to prevent data leakage (in CT-RATE). All projection angles are defined in an object-centered frame: $\theta = 0^\circ$ corresponds to the frontal (anteroposterior, AP) view, with the X-ray source in front of the object, and an increasing $\theta$ implies a counterclockwise rotation when viewed from above. These cases are split 7:1 for training and validation. Evaluations are done on a separate left-out test set of 50 cases. It is noted that optimization-based methods are directly optimized on the input projections at inference for testing. In comparison, our method operates in the projection domain and requires only a single projection per training sample, while learning-based methods require original ground-truth 3D volumes (CT-RATE) or volumes reconstructed from all available projections via FBP algorithm~\cite{shepp1974fourier} (SinoCT). This fundamental difference in training data requirements motivates our radiation reduction argument at the method level.

\begin{table*}[t]
    \centering
    \caption{Projection geometry and dataset splits (training, validation and testing) for CT-RATE and SinoCT. Our method uses only a single 2D projection per training sample, while learning-based baselines require the full 3D volumes reconstructed from dense angular sampling.}
    \label{tab:ds}
    \resizebox{\linewidth}{!}{
    \begin{tabular}{ccccccc}
        \toprule
       \multirow{2}{*}{Dataset} &\multicolumn{4}{c}{Projection Geometry}  & No. Training & No. Validation  \\
       \cmidrule(lr){2-5}
       & Detector Shape& Volume Shape& Detector Pixel Size&Voxel Size& Cases & Cases  \\
       \midrule
        CT-RATE~\cite{hamamci2024developing}&$256\times256$&$256\times256\times256$&$2\times2$&$2\times2\times2$&4200&600\\
        SinoCT~\cite{hooper2021impact}&$64\times512$&$64\times512\times512$&$4\times1$&$4\times1\times1$&2800&400\\
        \bottomrule
    \end{tabular}}
\end{table*}

\noindent\textbf{Implementation Details}
We implemented our diffusion model for smooth reproduction based on Stable Diffusion~\cite{rombach2021highresolution}, which uses a U-Net~\cite{ronneberger2015u} backbone interleaved with cross-attention layers. We set the model channels of our U-Net to be [64, 128, 256, 512, 1024] and apply pose injection at all encoder and decoder layers except the first and second U-Net layer for efficient computation, with a pose context dimension of 64. The final model has 140M parameters.

\noindent\textbf{Training protocol}
We use AdamW~\cite{loshchilov2017decoupled} for diffusion model training with a constant learning rate of $4\times10^{-5}$, momentum 0.99, weight decay $10^{-5}$, and batch size 8 per GPU on 4$\times$ NVIDIA RTX Ada 6000 GPUs. This diffusion model was trained for 300 epochs (which takes approximately 5 days) on our GPU cluster. 

\begin{figure}
\vspace{-10pt}
    \centering
    \includegraphics[width=0.7\linewidth]{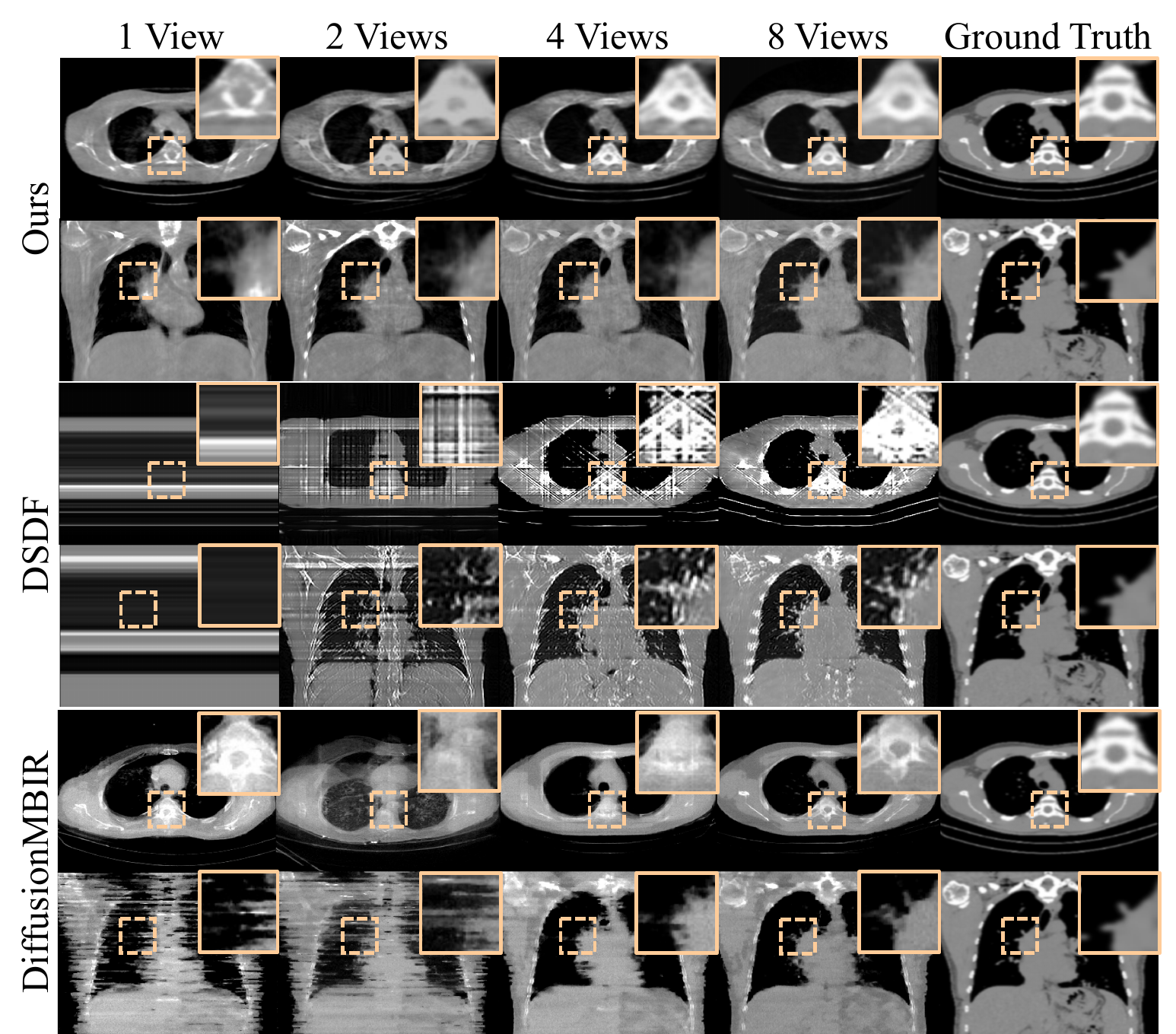}
    \caption{Tomographic reconstruction across different xSV-CT settings comparing our method against two best-performing methods~\cite{pan2025draw,chung2023solving}. One axial and one coronal slice is shown for each case.
    Our method produces more plausible reconstructions with fewer artifacts across all sparsity levels.
    }
    \label{fig:res2}

\end{figure}

\noindent\textbf{Inference hyperparameters}
For inference, we use moment order $n{=}20$ for the Range Condition (Eq.~\ref{alpha_z} in the main text), loss weights $\lambda_1{=}10^{-2}$ and $\lambda_2{=}1$, and generate $N{=}180$ novel projections.
Our method does \emph{not} require retraining when the number of input views $M$ changes at inference, unlike several learning-based methods (\eg, DSDF). The same trained model handles 1-, 4-, and 8-view inputs (as well as other view counts in Fig.~\ref{fig:res2} and Fig.~\ref{fig:moreviews} hereafter), with the consistency losses adapting automatically to the available measurements.

\noindent\textbf{Reconstruction cost} Our model is trained only once per dataset and can be applied at inference with different numbers of input views without retraining. Empirically, the inference time is dominated by the number of synthesized novel projections $N$, following an approximately linear trend of $t \approx 10+0.2N$ minutes. For reference, prior diffusion-based learning-based methods~\cite{song2024diffusionblend,chung2024decomposed} reports a inference time cost of around 40 minutes, this is comparable with approximately 46 minutes for our method in the 4-view setting ($N=180$). On CT-RATE with $256\times256$ projections, the peak VRAM usage is 4.1\,GB during training and 0.9\,GB during inference per batch.

\begin{table}[h]
  \centering
  \small
  \caption{Training paradigms of learning-based baselines and ours.}
  \label{tab:baselines}
  \begin{tabular}{lll}
    \toprule
    Method &  Supervision & Retraining \\
    \midrule
    DiffusionMBIR~\cite{chung2023solving} & Reconstructed axial slices & Per dataset (view-count agnostic) \\
    C$^2$RV~\cite{lin2024c} & Reconstructed 3D volumes & Per dataset and per view count \\
    DSDF~\cite{pan2025draw} & Reconstructed 3D volumes & Per dataset and per view count (both stages) \\
    Ours & X-ray projections & Per dataset (view-count agnostic) \\
    \bottomrule
  \end{tabular}
\end{table}
\paragraph{Comparison methods} Our method is trained on a single projection at a randomly sampled pose per sample and accepts inputs at arbitrary poses at inference. To evaluate all methods under this setting, we use the official implementation of each comparison method and adapt it to our training setup where needed. Optimization-based methods need no adaptation, since they are optimized directly on the input projections of each test case. DiffusionMBIR needs none either: its prior is trained on axial slices, and the acquisition geometry enters only through the forward operator at inference, so we follow its original training setup. We retrain C$^2$RV and DSDF separately for each view count $M$, drawing the $M$ input projections at randomly sampled poses in every training iteration.  All learning-based methods use the same splits and preprocessing as our method, with geometry and resolution matched to ours, and are trained until the loss stops decreasing for 10 epochs. Table~6 summarizes their supervision and retraining scope.

\section{Extended Analysis}
\label{sec:C}

\begin{figure}[t]
    \centering
    \includegraphics[width=\linewidth]{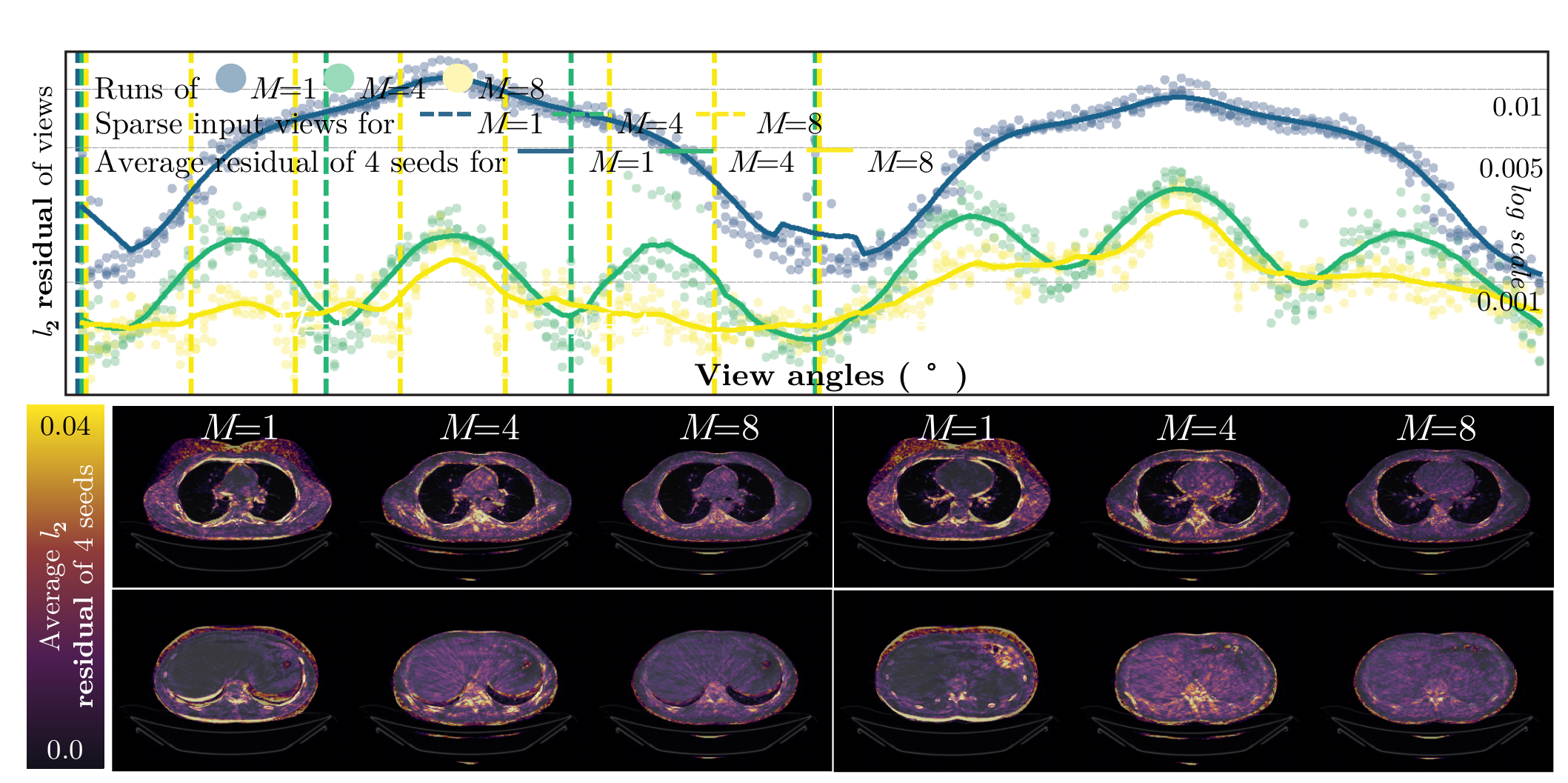}
    \caption{Consistency analysis via $\ell_2$ residuals under the 3$\times$SV-CT setting across 4 random seeds. \textbf{Top:} residuals of novel view synthesis at different angles, showing reduced error near input angles and consistent behavior across seeds. \textbf{Bottom:} average residuals of tomographic reconstruction overlayed on ground-truth axial slices, showing that reconstruction error decreases as more views become available. The low cross-seed variance suggests that the guided generation produces similar outputs across different initializations. }
    \label{fig:moreang}
\end{figure}

\begin{figure}[t]
    \centering
    \includegraphics[width=\linewidth]{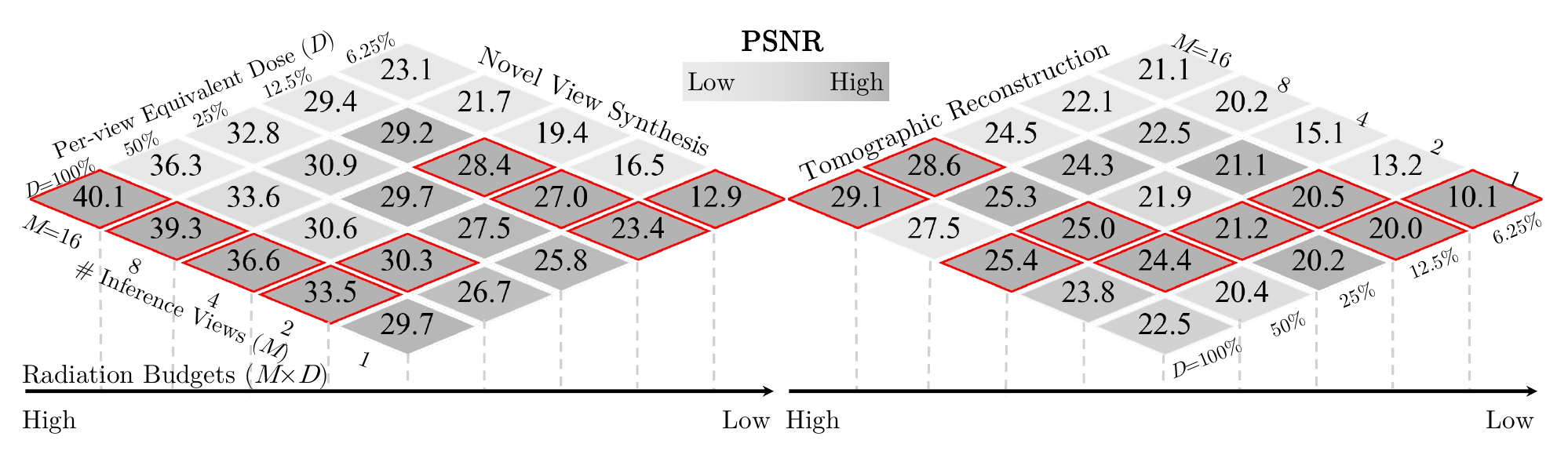}
    \caption{Performance under fixed radiation budgets: each vertically aligned group of cells corresponds to a constant radiation budget (inference views $M$ $\times$ per-view dose $D$). Red outlines indicate the best-performing combination of $M$ and $D$ under each budget. Numbers denote PSNR values for either Novel View Synthesis (left) or Tomographic Reconstruction (right) on CT-RATE.  The optimal operating point consistently involves moderate sparsity paired with moderate per-view dose, suggesting that our framework could provide flexibility in balancing view count and per-view exposure under radiation constraints.}
    \label{fig:sweep}
\end{figure}

\begin{table}[h]
\centering
\setlength{\tabcolsep}{5mm}
\caption{Effect of training-time data availability: PSNR for novel view synthesis / tomographic reconstruction on CT-RATE dataset with more views per sample seen at training. With a single projection per training sample, our method achieves competitive results. Additional training views provide gains, particularly for single-view inference, while multi-view inference is less sensitive to the number of views per training sample.}
\label{tab:moreviews}
\resizebox{0.7\linewidth}{!}{
\begin{tabular}{c ccc}
\toprule
\multirow{2}{*}{\#Train Views/Sample} & \multicolumn{3}{c}{\#Test Views} \\
\cmidrule(lr){2-4}
 & 1 & 4 & 8 \\
\midrule
1 (Ours)  & 23.1 / 21.1 & 32.8 / 24.5 & 36.3 / 28.6 \\
4  & 23.0 / 21.1 & 33.2 / 24.8 & 36.9 / 28.6 \\
32 & 26.0 / 21.3 & 33.6 / 24.8 & 38.1 / 29.3 \\
\bottomrule
\end{tabular}}
\end{table}

\begin{figure}[t]
    \centering
    \includegraphics[width=0.7\linewidth]{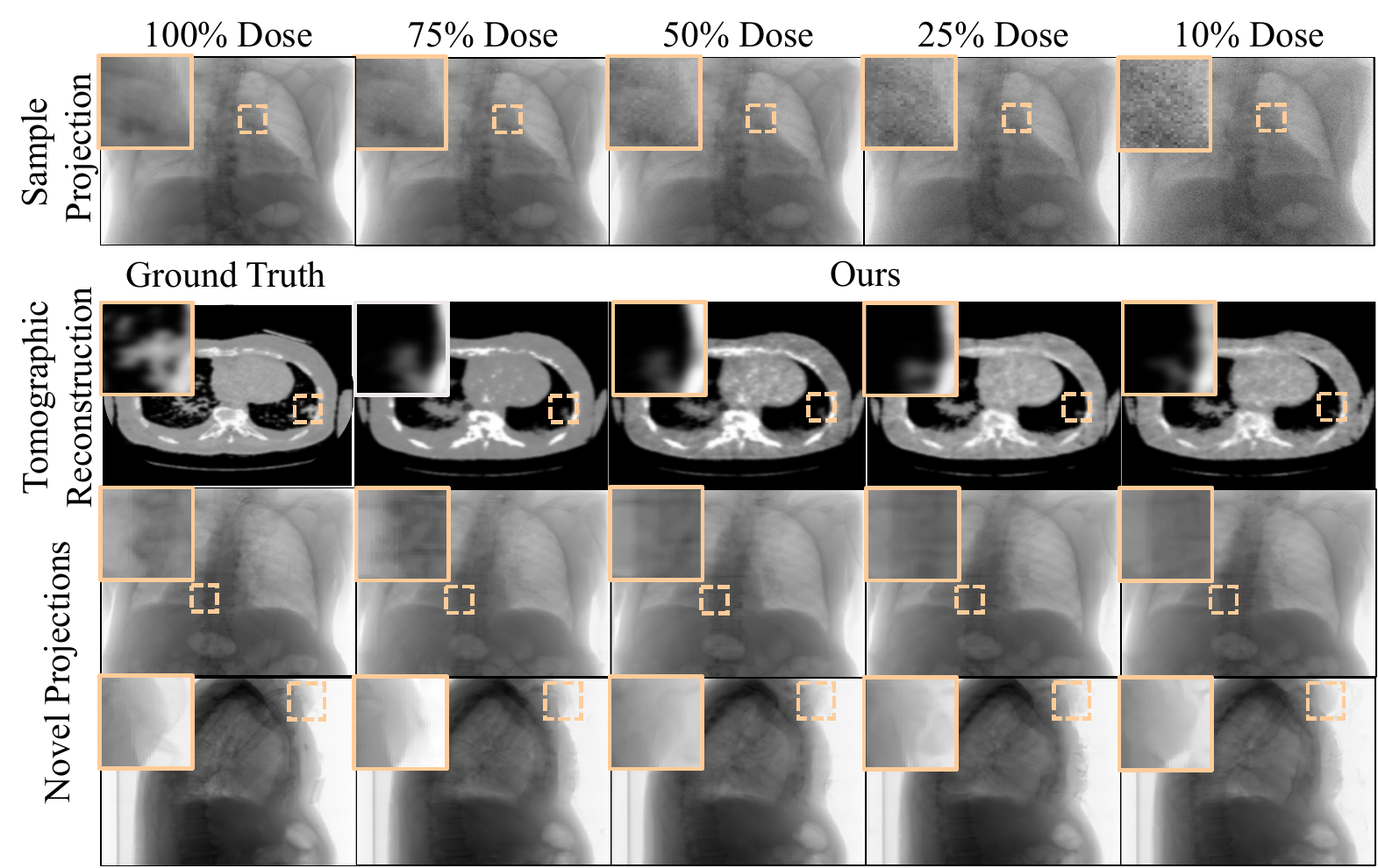}
    \caption{Qualitative evaluation under noisy input projections. Each row shows reconstruction from 8 noisy projections (one example shown in the first row) under simulated 75\%, 50\%, 25\% and 10\% of original X-ray dosage to guide the inference process.}
    \label{fig:robust}
\end{figure}
% \begin{table}[t]
% \centering
% \footnotesize
% \setlength{\heavyrulewidth}{0.11em}
% \begin{minipage}[t]{0.49\linewidth}
%     \centering
%     \setlength{\tabcolsep}{1.mm}
%     \caption{Robustness to noisy input projections. Metrics (\textbf{P}SNR/\textbf{S}SIM/\textbf{L}PIPS) are averaged on 50 CT-RATE cases with 8 input views.}
%     \label{tab:robust1}
%     \resizebox{\linewidth}{!}{
%     \input{tabs/tab_robust1}
%     }
% \end{minipage}
% \begin{minipage}[t]{0.49\linewidth}
%     \centering
%     \setlength{\tabcolsep}{1.mm}
%     \caption{Robustness to angular uncertainty. Metrics are averaged on 50 CT-RATE cases with 8 input views.}
%     \label{tab:robust2}
%     \resizebox{\linewidth}{!}{
%     \input{tabs/tab-robust2}}
% \end{minipage}
% \end{table}
% \vspace{-10pt}

\begin{table}[t]
    \centering
    \scriptsize
    \setlength{\tabcolsep}{4pt}
    \caption{Robustness to noisy input projections (left) and angular uncertainty (right). Metrics (\textbf{P}SNR/\textbf{S}SIM/\textbf{L}PIPS) are averaged on 50 CT-RATE cases with 8 input views.}
    \label{tab:robust}
    \resizebox{0.44\linewidth}{!}{
\begin{tabular}{c ccc ccc}
    \toprule
       \multirow{1}{*}[-0.35ex]{ Equiv.} &\multicolumn{3}{c}{Novel View Syn.} & \multicolumn{3}{c}{Reconstruction} \\
        \cmidrule(lr){2-4}\cmidrule(lr){5-7}
        \multirow{1}{*}[.35ex]{ Dosage}&\textbf{P}$\uparrow$&\textbf{S}$\uparrow$&\textbf{L}$\downarrow$&\textbf{P}$\uparrow$&\textbf{S}$\uparrow$&\textbf{L}$\downarrow$\\
        \midrule
        100\%&36.3&0.95&0.17&28.6&0.69&0.27\\
        \cmidrule(lr){1-7}
        75\%&34.3&0.90&0.20&26.1&0.66&0.28\\
        50\%&33.6&0.89&0.23&25.3&0.66&0.30\\
        25\%&30.6&0.85&0.24&25.0&0.65&0.31\\
        10\%&31.2&0.85&0.24&24.3&0.61&0.30\\
    \bottomrule
    \end{tabular}}
    \hspace{5pt}
    \resizebox{0.43\linewidth}{!}{
    \begin{tabular}{c ccc ccc}
    \toprule
       \multirow{2}{*}[-0.5ex]{$\Delta\theta$} &\multicolumn{3}{c}{Novel View Syn.} & \multicolumn{3}{c}{Reconstruction} \\
       \cmidrule(lr){2-4}\cmidrule(lr){5-7}
         &\textbf{P}$\uparrow$&\textbf{S}$\uparrow$&\textbf{L}$\downarrow$&\textbf{P}$\uparrow$&\textbf{S}$\uparrow$&\textbf{L}$\downarrow$\\
        \midrule
        $\pm0^\circ$&36.3&0.95&0.17&28.6&0.69&0.27\\
        \cmidrule(lr){1-7}
        $\pm2.5^\circ$&34.4&0.93&0.21&27.3&0.68&0.33\\
        $\pm5^\circ$&34.5&0.94&0.20&24.9&0.66&0.35\\
        $\pm7.5^\circ$&32.8&0.88&0.27&21.6&0.57&0.39\\
        $\pm10^\circ$&29.1&0.82&0.35&20.6&0.54&0.39\\
    \bottomrule
    \end{tabular}}
    \vspace{-10pt}
\end{table}

\noindent\textbf{Generation Faithfulness and Consistency}
A critical question for any generative approach in medical imaging is the extent to which generated content is constrained by measurements versus injected by the learned prior. Under extreme sparsity (1--4 views), the problem is severely ill-posed and much of the reconstruction is necessarily prior-driven. In the main text, we evaluate the measurement-dependence of our generation quantitatively in Table~\ref{tab:ab1}, where removing $\mathcal{L}_{\rm geo}$ causes a large performance drop. The measurement fidelity term in $\mathcal{L}_{\rm geo}$ (Eq.~\ref{eq:lgeo}) constrains generated views to be compatible with the actual X-ray measurements, not just anatomically plausible. We emphasize that measurement fidelity at a few input angles does not guarantee anatomical correctness at all generated angles, since structures far from the measured views remain substantially prior-driven.
To evaluate the consistency, we run our method 4 times with different random seeds under the 3$\times$SV-CT setting and report the $\ell_2$ residual of (i) generated novel projections at different angles (Fig.~\ref{fig:moreang}, top), and 
(ii) reconstructed CT slices (Fig.~\ref{fig:moreang}, bottom).
The model behaves consistently across seeds: generated projections at angles closer to the sparse input views retain more structural detail than those with larger angular separation, and this effect diminishes as the number of available projections $M$ at inference increases. The low cross-seed variance suggests that the guided generation produces repeatable outputs rather than arbitrary samples from the prior. This consistent angular pattern also makes it possible to estimate a per-view confidence map without ground truth, from the variance of generated projections across random seeds. Even a simple estimate based on the 4 seeds above correlates with the per-view $\ell_2$ error (Pearson $r = 0.60$ for $M=4$), suggesting that cross-seed variance can flag generated views that rely more heavily on the learned prior to a certain extent.

\noindent\textbf{Radiation Budget Trade-off}
In clinical CT, the total radiation budget $B = M \times D$ (number of views $\times$ per-view dose) is the primary safety constraint. Given a fixed budget, should one acquire few high-dose views or many low-dose views?
We evaluate this trade-off under simulated fixed radiation budgets in Fig.~\ref{fig:sweep} by sweeping over inference views $M$ and per-view dose $D$ under fixed radiation budgets. The results show that the model performs best under a moderate combination of view sparsity and measurement noise for both novel view synthesis and tomographic reconstruction tasks. This observation highlights the importance of developing xSV-CT frameworks that provide flexibility in balancing projection sparsity and noise levels when operating under a fixed radiation budget in clinical settings~\cite{gao2014low}.

\noindent\textbf{Training-time Scalability} We further evaluate the scalability of our model when trained with increasing numbers of projections per sample. As shown in Tab.~\ref{tab:moreviews}, performance improves gradually as more views are available during training, indicating that our model can effectively leverage additional training projections and scale with increased data availability.

\noindent\textbf{Model Robustness} Inspired by the model self-calibration analysis in NeAT~\cite{ruckert2022neat}, we evaluate the robustness of our method under two practical perturbation settings: (\textbf{1}) Robustness to noisy projections: In real-world acquisition, even with nominally accurate geometric calibration, projection quality can be degraded by changes in the imaging environment, reduced photon counts, and electronic readout noise. This issue is amplified in xSV-CT, where the limited number of projections provides fewer measurements over which noise can be averaged out. Following the low-dose X-ray noise simulation protocols in~\cite{yu2012development,vzabic2013low}, we use TIGRE~\cite{biguri2016tigre} to inject mixed Poisson and Gaussian noise into the input views, modeling photon-counting noise and detector/electronic noise under reduced-dose acquisition. As shown in Table~\ref{tab:robust} (left) and Fig.~\ref{fig:robust}, our method degrades gracefully under noisy inputs: the global anatomical structure remains stable, while most degradation is confined to fine-scale details. This behavior suggests that the diffusion prior provides a useful structural constraint when individual sparse projections become unreliable. (\textbf{2}) Robustness to uncertainty of camera poses: We further evaluate robustness to geometric miscalibration, where the assumed acquisition angles deviate from the true scanning geometry. Specifically, we perturb each input angle $\theta\in\boldsymbol{\theta}$ by a random offset sampled within a prescribed range $\Delta\theta$. Table~\ref{tab:robust} (right) shows that our method outperforms competing approaches when the angular error is within $\pm2.5^\circ$ and remains competitive up to $\pm7.5^\circ$, indicating tolerance to moderate pose-calibration errors. Fig.~\ref{fig:analysis} provides visual comparisons under both noisy projections and angular perturbations. Compared with selected baselines, our method better preserves coherent anatomical structures and fine details under both types of acquisition uncertainty.

\section{Failure Analysis}
\label{sec:D}
\begin{figure*}[h!]
    \centering
    \includegraphics[width=\linewidth]{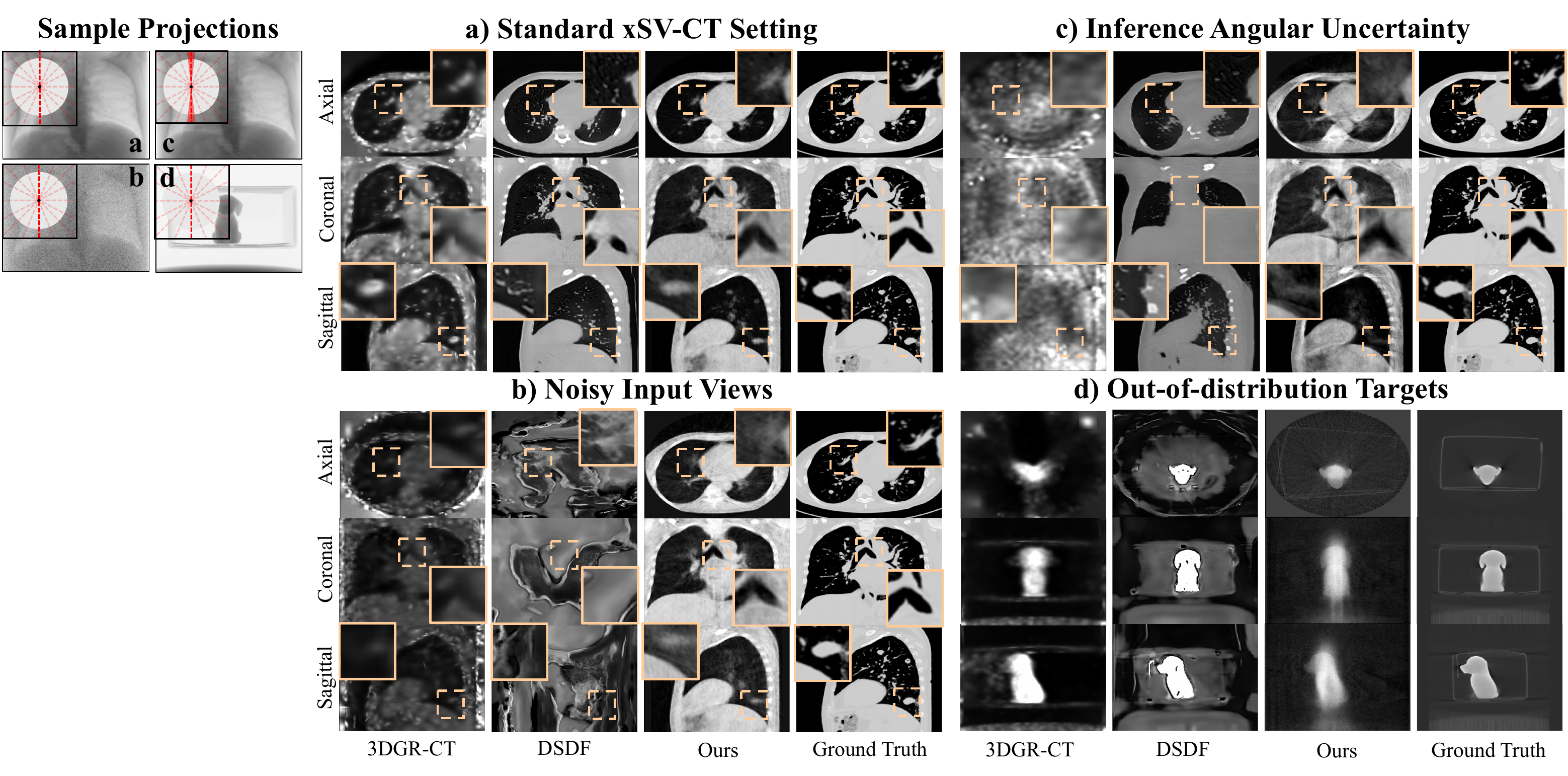}
    \caption{Systematic analysis of failure modes across four conditions, comparing our method with the best-performing optimization-based (3DGR-CT~\cite{li20253dgr}) and learning-based methods (DSDF~\cite{pan2025draw}) using 8 input views (0$^\circ$, 22.5$^\circ$, 45$^\circ$, 67.5$^\circ$, 90$^\circ$, 112.5$^\circ$, 135$^\circ$, 157.5$^\circ$). The model of ours and the learning-based method are trained on CT-RATE dataset. Sample input projections as well as input angles are shown on the left. We provide tomographic reconstruction results under 4 different settings: 
    \textbf{a}) Standard xSV-CT setting with 8 input views:  our method preserves fine-grained anatomical details. 
    \textbf{b}) xSV-CT with a simulated 5\% dose of the 8 input views: our diffusion prior maintains structural integrity while baselines introduce severe artifacts. 
    \textbf{c}) xSV-CT with the 8 input view angles randomly perturbed within $\pm10^\circ$: our method stays anchored to the available projections, while baselines develop geometric artifacts (3DGR-CT) or hallucinate structure (DSDF).
    \textbf{d}) xSV-CT on an out-of-distribution (OOD) target with respect to the training datasets: our method exhibits graceful blurring along the superior-inferior axis, yet still produces more faithful reconstructions than both baselines (note that for optimization-based methods there are no OOD targets in this sense).}
    \label{fig:analysis}
\end{figure*}

Understanding failure modes is essential for responsible deployment. While our method demonstrates robustness to noisy projections and angular uncertainties (Table~\ref{tab:robust}), we characterize the conditions under which it exhibits graceful degradations. For example, as shown in Table~\ref{tab:robust}, the performance metrics drop when the input X-ray projections have excessive noise due to low dosage outputs (\eg, less than a simulated 10\% of that in the normal-dose CT dataset it is trained on) or if the angular input uncertainty exceeds a certain threshold (\eg, more than a $\pm7.5^\circ$ of error between the model's angular input and the true camera angular pose). 
We visualize these failure modes in Fig.~\ref{fig:analysis}(\textbf{b},\textbf{c}), with reconstructions from ours and the best-performing optimization-based (3DGR-CT~\cite{li20253dgr}) and compared learning-based (DSDF~\cite{pan2025draw}) methods. For context, Fig.~\ref{fig:analysis}(\textbf{a}) shows results under the standard xSV-CT setting, where our method excels at preserving fine-grained details.

\noindent\textbf{Input X-ray dosage}
When the input X-ray dosage drops below $\sim$10\% of the normal-dose level, the signal-to-noise ratio of the input projections deteriorates to the point where $\mathcal{L}_{\rm geo}$'s measurement fidelity term anchors the generation to noisy observations. Nonetheless, the diffusion prior maintains the global structural integrity of the reconstruction. Our method degrades more gracefully than both baseline methods, which introduce severe artifacts. 

\noindent\textbf{Angular uncertainty}
When the angular input error exceeds $\pm7.5^\circ$, the mismatch between the assumed and true acquisition angles causes the geometric coherency loss to enforce consistency with an incorrect geometry. This manifests as local geometric distortions. Compared methods either develop geometric artifacts~\cite{li20253dgr} or hallucinate structure~\cite{pan2025draw}, whereas our method stays anchored to the available projections and recovers the anatomy more faithfully.

\noindent\textbf{Out-of-distribution Targets}
Like the learning-based methods, our approach's performance is impaired when applied to out-of-distribution (OOD) targets, \textit{i.e.}, data that lies outside the training distribution. As visualized in Fig.~\ref{fig:analysis}(\textbf{d}), OOD inputs introduce noticeable blurring effects along the superior-inferior axis of the target and slight distortions at anatomical boundaries with our method. This is expected: the learned prior cannot synthesize anatomy it has never observed. However, even under these OOD conditions, our model still produces more visually faithful reconstructions compared to both the optimization-based~\cite{li20253dgr} and learning-based~\cite{pan2025draw} compared methods, thanks to the physics-based consistency constraints that remain valid regardless of anatomy. 

The identified failure modes suggest concrete directions for future work: (1) noise-aware training or adaptive guidance weighting for low-dose robustness, (2) joint angle-projection optimization for angular uncertainty, and (3) foundation model pre-training or few-shot adaptation for OOD generalization.

\textbf{Robustness to off-model physics} To stress-test structured violations of the Radon model, we simulate metal-induced beam hardening following~\cite{zhang2018convolutional}. We insert several ellipsoidal metal implants (about 4000 voxels each, with intensities of 3000 HU under standard FBP reconstruction) into 10 CT-RATE cases and compute PSNR against the clean, metal-free volume (Table~\ref{tab:metal}). FBP from 360 corrupted views propagates beam-hardening streaks throughout the volume, whereas our 8-view reconstruction retains fewer artifacts (26.4 vs.\ 20.7\,dB). Outside the metal region, our method reaches 27.7\,dB, only 0.9\,dB below reconstruction from clean projections, while FBP remains at 21.2\,dB.
 
\begin{table}[h]
  \centering
  \small
  \caption{Reconstruction under simulated metal-induced beam hardening. PSNR is computed against the clean, metal-free volume; values in parentheses give the change relative to reconstructing from clean projections with our methods with $M=8$.}
  \label{tab:metal}
  \begin{tabular}{lc}
    \toprule
    Variant & PSNR$\uparrow$ \\
    \midrule
    FBP, $M{=}360$, corrupted projections & 20.7 \\
    FBP, $M{=}360$, corrupted projections, non-metal region & 21.2 \\
    Ours, $M{=}8$, corrupted projections & 26.4 ($-2.2$) \\
    Ours, $M{=}8$, corrupted projections, non-metal region & 27.7 ($-0.9$) \\
    \bottomrule
  \end{tabular}
\end{table}

\section{Comprehensive Qualitative Results}
\label{sec:E}
% \noindent\textbf{Tomography and Novel Projection Synthesis}
We present more qualitative analysis for tomographic reconstruction and novel view synthesis across CT-RATE and SinoCT datasets under three challenging xSV-CT settings (1, 4, and 8 views). Two representative cases from each dataset are detailed in Table~\ref{tab:link}, which provides an index to the corresponding visualizations listed in this appendix. These comparisons show that our method achieves superior quantitative performance (Tables~\ref{tab:compare1} and~\ref{tab:compare2}) and produces more visually plausible results. Please zoom in to compare fine anatomical details across methods. For further qualitative 3D validation, we refer to the accompanying \textbf{video}, which visualizes these cases across compared methods and ours. 

\begin{table}[hb]
    \centering
    \caption{Index of qualitative results. Each entry links to the corresponding figure showing novel view synthesis (NVS) or tomographic reconstruction (TR) for a specific case and sparsity level.}
    \label{tab:link}
    \resizebox{0.7\linewidth}{!}{
    \begin{tabular}{cc ccc ccc}
    \toprule
        \multicolumn{2}{c}{Case No.} & \multicolumn{3}{c}{Case 1}& \multicolumn{3}{c}{Case 2}\\
        \cmidrule(lr){1-2} \cmidrule(lr){3-5} \cmidrule(lr){6-8}
        \multicolumn{2}{c}{No. Inference Views} & 1-view & 4-view & 8-view &  1-view & 4-view & 8-view \\
        \midrule
        \multirow{2}{*}{CT-RATE} & NVS & Fig.~\ref{fig:c1-1} & Fig.~\ref{fig:c1-4} & Fig.~\ref{fig:c1-8} & Fig.~\ref{fig:c2-1} & Fig.~\ref{fig:c2-4} & Fig.~\ref{fig:c2-8} \\
         & TR &Fig.~\ref{fig:c1-1-} & Fig.~\ref{fig:c1-4-} & Fig.~\ref{fig:c1-8-} & Fig.~\ref{fig:c2-1-} & Fig.~\ref{fig:c2-4-} & Fig.~\ref{fig:c2-8-} \\
        \cmidrule(lr){1-8}
        \multirow{2}{*}{SinoCT} & NVS & Fig.~\ref{fig:s1-1} & Fig.~\ref{fig:s1-4} & Fig.~\ref{fig:s1-8} & Fig.~\ref{fig:s2-1} & Fig.~\ref{fig:s2-4} & Fig.~\ref{fig:s2-8} \\
         & TR &Fig.~\ref{fig:s1-1-} &Fig.~\ref{fig:s1-4-} & Fig.~\ref{fig:s1-8-} & Fig.~\ref{fig:s2-1-} & Fig.~\ref{fig:s2-4-} & Fig.~\ref{fig:s2-8-} \\
        \bottomrule
    \end{tabular}}
\end{table}

\section{Broader Impacts}
\label{sec:F}

Beyond the conventional motivation of sparse-view reconstruction, \textit{i.e.}, reducing radiation exposure and associated health risks for both patients and operators, xSV-CT has the potential to broaden the range of scenarios in which tomographic reconstruction is practical. In intraoperative navigation, for example, current workflows often rely directly on planar X-ray projections for guidance; an xSV-CT regime could provide volumetric information from only a few acquisitions. For dynamic imaging scenarios, such as cardiac imaging, or high-throughput industrial inspection, reducing the number of required projections can shorten acquisition time, thereby improving temporal resolution or productivity. For radiation-sensitive biological samples, cultural-heritage artifacts, or archaeological specimens, the reduced imaging dose may also enable tomographic analysis of objects that would otherwise be difficult or unsafe to scan repeatedly.

However, these potential benefits should also be interpreted with caution. Like other sparse-view CT methods, xSV-CT operates under severely limited measurements and can therefore introduce reconstruction bias, hallucinated structures, or degradation of fine details. Such risks are particularly important in safety-critical clinical or industrial settings, where reconstructed volumes should not be treated as definitive evidence without appropriate validation, uncertainty assessment, and expert oversight. In addition, the current computational cost makes our method more suitable as an asynchronous reconstruction module: a full reconstruction may still require tens of minutes. Compared with analytical reconstruction methods that can produce results within minutes or less, this latency may limit applicability in real-time workflows. Reducing inference cost and quantifying reconstruction uncertainty remain important directions for future deployment.

% \bibliographystyle{splncs04}
% \bibliography{main}
\clearpage
\begin{figure*}[t]
    \centering
    \includegraphics[width=1\linewidth]{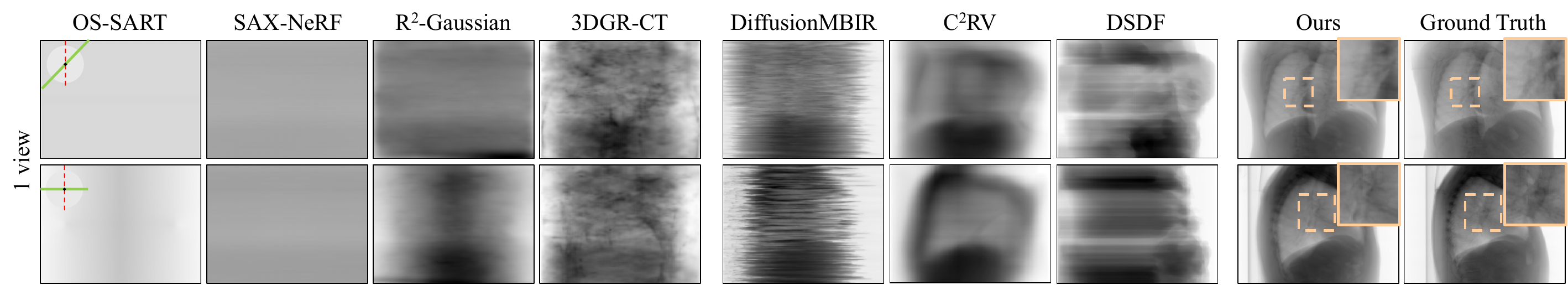}
    \caption{Novel View Synthesis results (45$^\circ$, 90$^\circ$) from 1 view (0$^\circ$) on Case 1 from the CT-RATE dataset, the input (red) and novel (green) angles are also indicated in the inset polar graph. We provide zoomed-in details for results with plausible anatomy compared to their corresponding ground truths.}
    \label{fig:c1-1}
\end{figure*}
\begin{figure*}[t]
    \centering
    \includegraphics[width=1\linewidth]{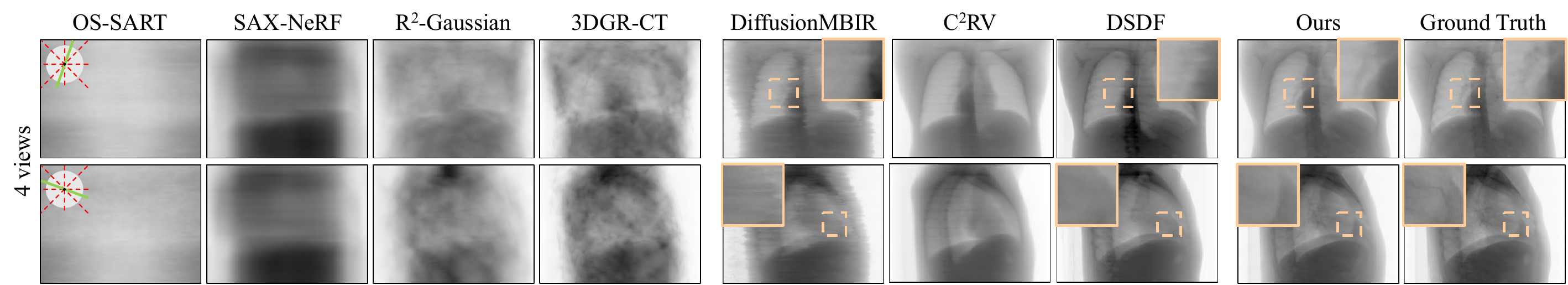}
    \caption{Novel View Synthesis results (22.5$^\circ$, 112.5$^\circ$) from 4 views (0$^\circ$, 45$^\circ$, 90$^\circ$, 135$^\circ$) on Case 1 from the CT-RATE dataset, the input (red) and novel (green) angles are also indicated in the inset polar graph. We provide zoomed-in details for results with plausible anatomy compared to their corresponding ground truths.}
    \label{fig:c1-4}
\end{figure*}
\begin{figure*}[t]
    \centering
    \includegraphics[width=1\linewidth]{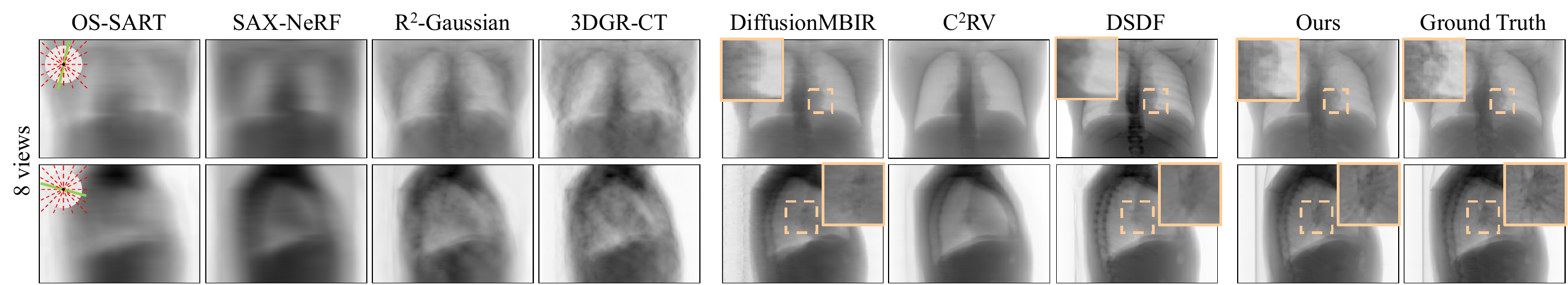}
    \caption{Novel View Synthesis results (11.25$^\circ$, 101.25$^\circ$) from 8 views (0$^\circ$, 22.5$^\circ$, 45$^\circ$, 67.5$^\circ$, 90$^\circ$, 112.5$^\circ$, 135$^\circ$, 157.5$^\circ$) on Case 1 from the CT-RATE dataset, the input (red) and novel (green) angles are also indicated in the inset polar graph. We provide zoomed-in details for results with plausible anatomy compared to their corresponding ground truths.}
    \label{fig:c1-8}
\end{figure*}
\begin{figure*}[t]
    \centering
    \includegraphics[width=1\linewidth]{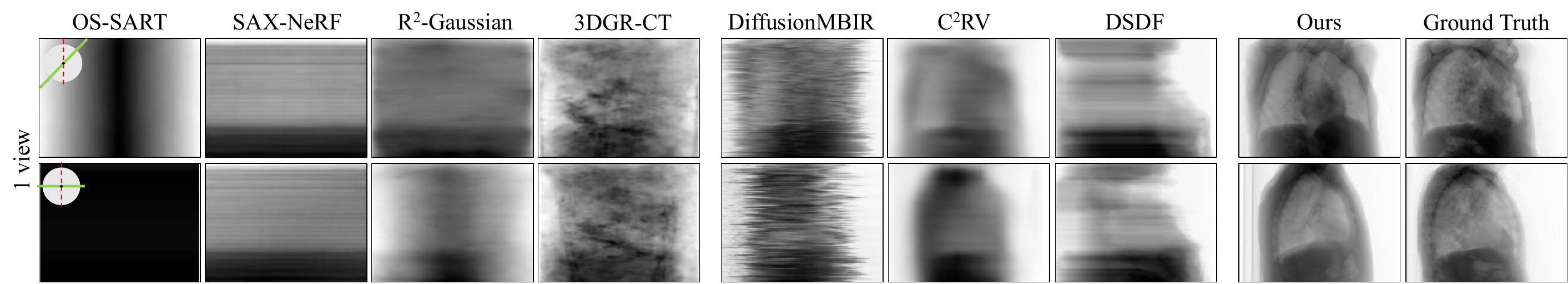}
    \caption{Novel View Synthesis results (45$^\circ$, 90$^\circ$) from 1 view (0$^\circ$) on Case 2 from the CT-RATE dataset, the input (red) and novel (green) angles are also indicated in the inset polar graph.}
    \label{fig:c2-1}
\end{figure*}
\begin{figure*}[t]
    \centering
    \includegraphics[width=1\linewidth]{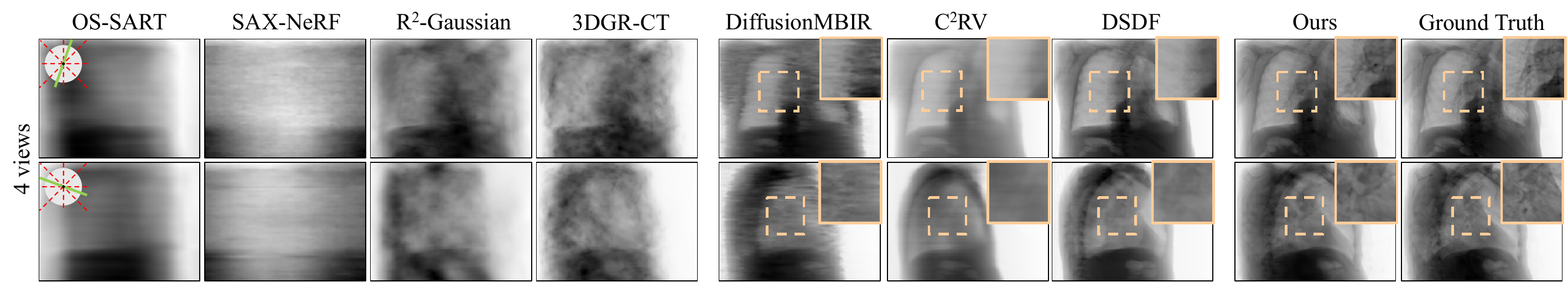}
    \caption{Novel View Synthesis results (22.5$^\circ$, 112.5$^\circ$) from 4 views (0$^\circ$, 45$^\circ$, 90$^\circ$, 135$^\circ$) on Case 2 from the CT-RATE dataset, the input (red) and novel (green) angles are also indicated in the inset polar graph. We provide zoomed-in details for results with plausible anatomy compared to their corresponding ground truths.}
    \label{fig:c2-4}
\end{figure*}
\begin{figure*}[t]
    \centering
    \includegraphics[width=1\linewidth]{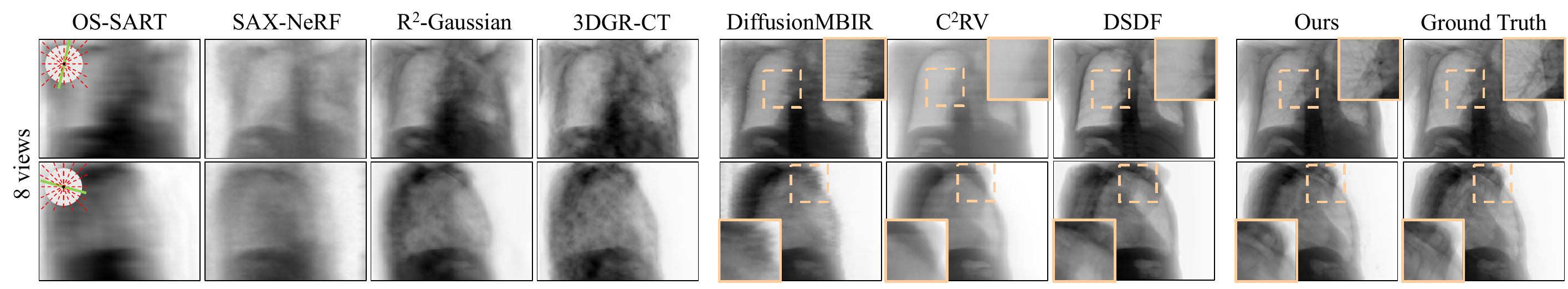}
    \caption{Novel View Synthesis results (11.25$^\circ$, 101.25$^\circ$) from 8 views (0$^\circ$, 22.5$^\circ$, 45$^\circ$, 67.5$^\circ$, 90$^\circ$, 112.5$^\circ$, 135$^\circ$, 157.5$^\circ$) on Case 2 from the CT-RATE dataset, the input (red) and novel (green) angles are also indicated in the inset polar graph. We zoom in on methods with plausible details with the ground truths.}
    \label{fig:c2-8}
\end{figure*}
\begin{figure*}[t]
    \centering
    \includegraphics[width=1\linewidth]{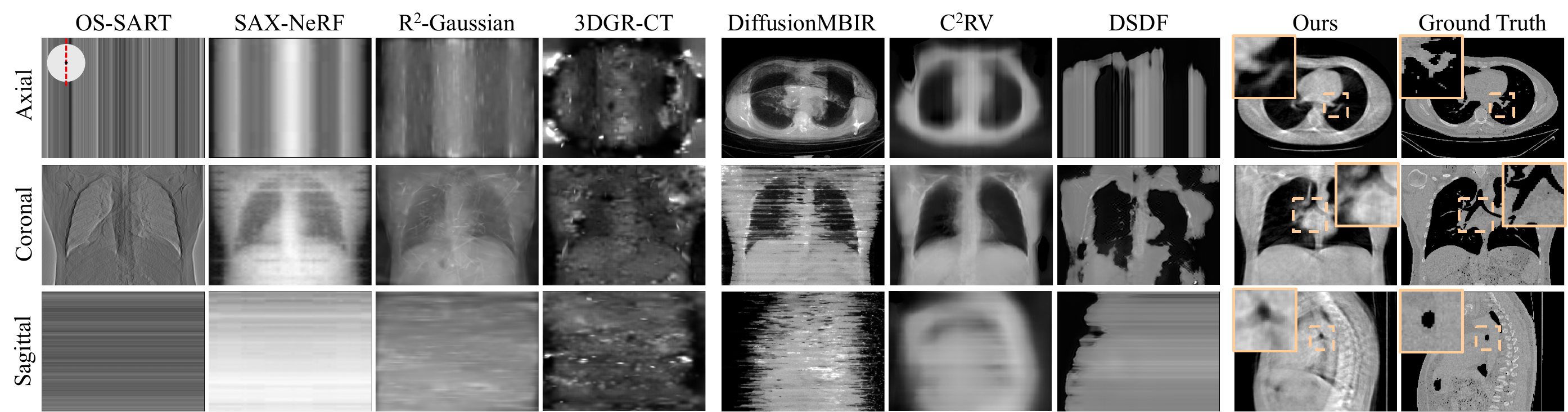}
    \caption{Tomographic Reconstruction results from 1 view (0$^\circ$) on Case 1 from the CT-RATE dataset, the input angle is also indicated in the inset polar graph. We zoom in on methods with plausible details with the ground truths.}
    \label{fig:c1-1-}
\end{figure*}
\begin{figure*}[t]
    \centering
    \includegraphics[width=1\linewidth]{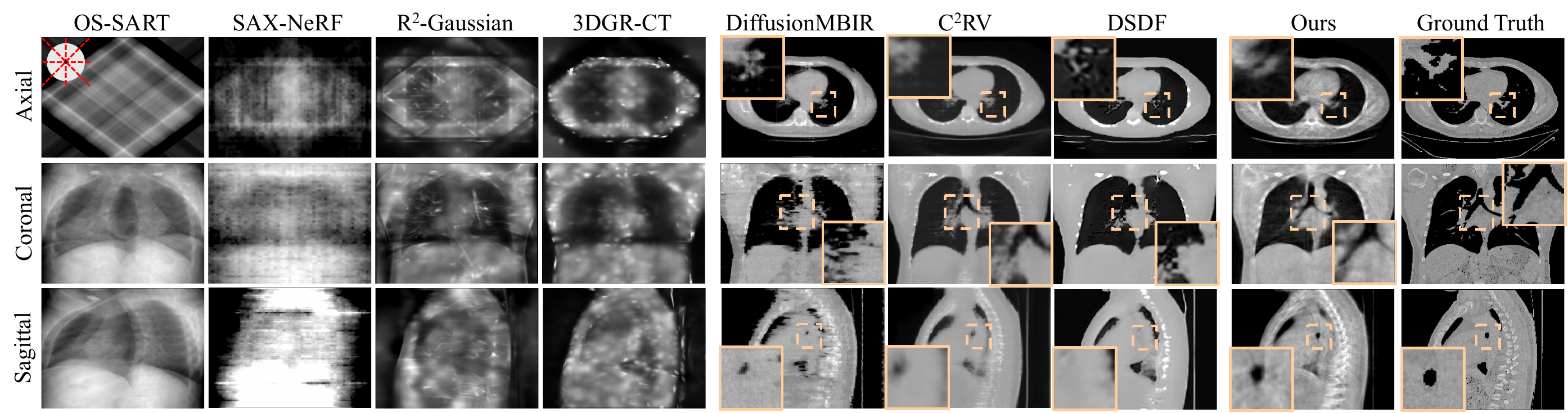}
    \caption{Tomographic Reconstruction results from 4 views (0$^\circ$, 45$^\circ$, 90$^\circ$, 135$^\circ$) on Case 1 from the CT-RATE dataset, the input angles is also indicated in the inset polar graph. We zoom in on methods with plausible details with the ground truths.}
    \label{fig:c1-4-}
\end{figure*}
\begin{figure*}[t]
    \centering
    \includegraphics[width=1\linewidth]{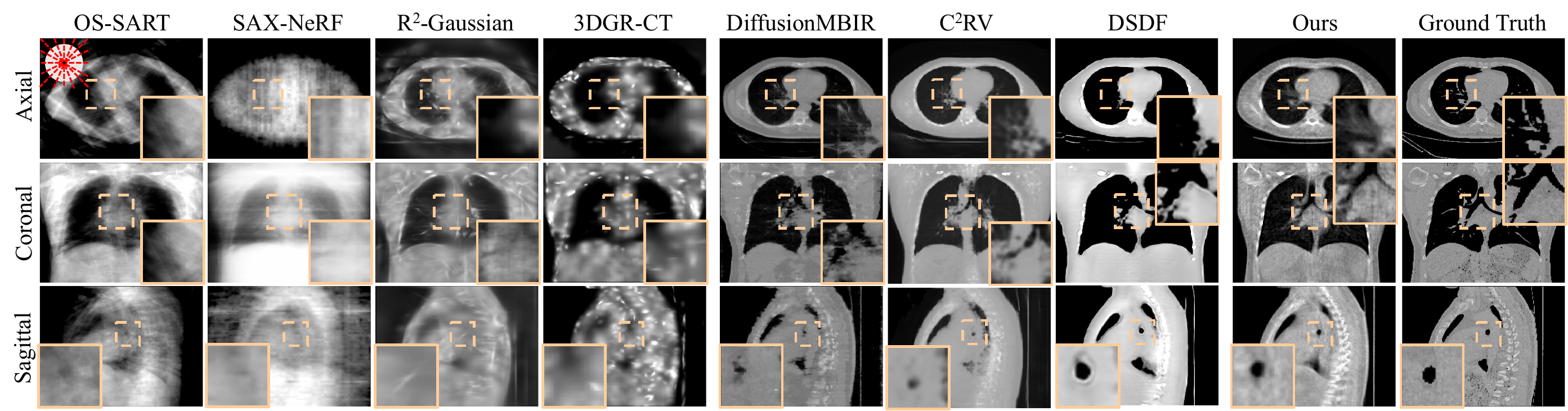}
    \caption{Tomographic Reconstruction results from 8 views (0$^\circ$, 22.5$^\circ$, 45$^\circ$, 67.5$^\circ$, 90$^\circ$, 112.5$^\circ$, 135$^\circ$, 157.5$^\circ$) on Case 1 from the CT-RATE dataset, the input angles is also indicated in the inset polar graph. We zoom in on methods with plausible details with the ground truths.}
    \label{fig:c1-8-}
\end{figure*}
\begin{figure*}[t]
    \centering
    \includegraphics[width=1\linewidth]{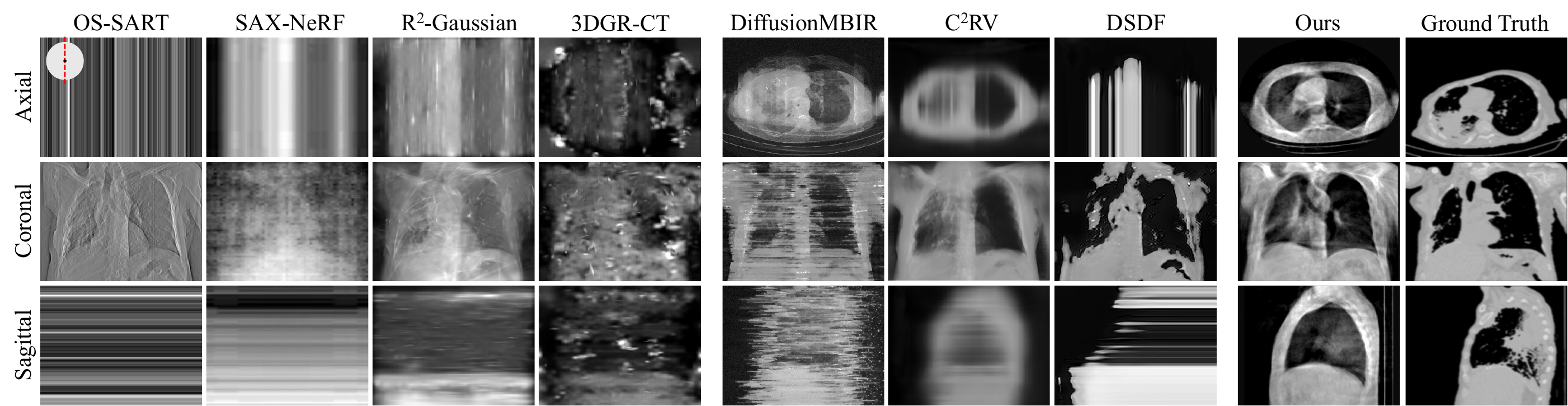}
    \caption{Tomographic Reconstruction results from 1 view (0$^\circ$) on Case 2 from the CT-RATE dataset, the input angle is also indicated in the inset polar graph. We zoom in on methods with plausible details with the ground truths.}
    \label{fig:c2-1-}
\end{figure*}
\begin{figure*}[t]
    \centering
    \includegraphics[width=1\linewidth]{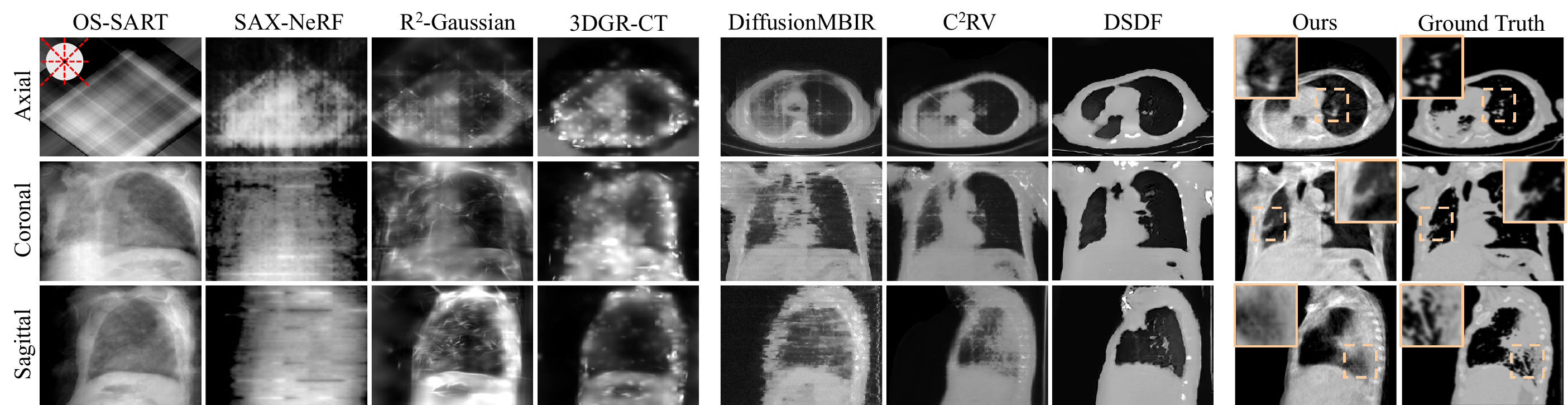}
    \caption{Tomographic Reconstruction results from 4 views (0$^\circ$, 45$^\circ$, 90$^\circ$, 135$^\circ$) on Case 2 from the CT-RATE dataset, the input angles is also indicated in the inset polar graph. We zoom in on methods with plausible details with the ground truths.}
    \label{fig:c2-4-}
\end{figure*}
\begin{figure*}[t]
    \centering
    \includegraphics[width=1\linewidth]{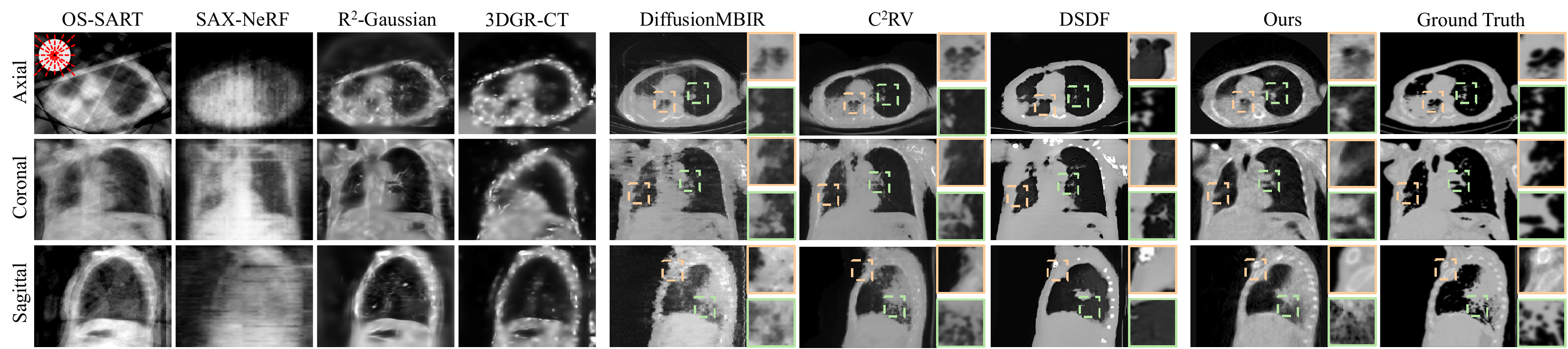}
    \caption{Tomographic Reconstruction results from 8 views (0$^\circ$, 22.5$^\circ$, 45$^\circ$, 67.5$^\circ$, 90$^\circ$, 112.5$^\circ$, 135$^\circ$, 157.5$^\circ$) on Case 2 from the CT-RATE dataset, the input angles is also indicated in the inset polar graph. We zoom in on methods with plausible details with the ground truths.}
    \label{fig:c2-8-}
\end{figure*}
\begin{figure*}[t]
    \centering
    \includegraphics[width=1\linewidth]{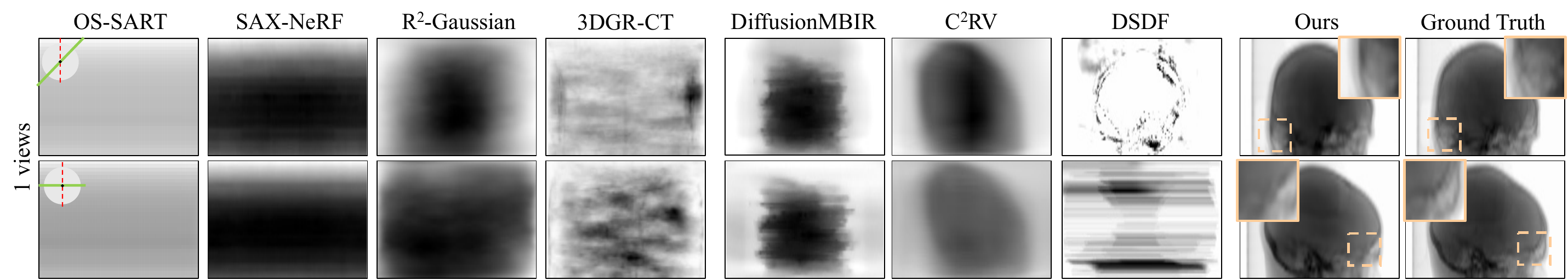}
    \caption{Novel View Synthesis results (45$^\circ$, 90$^\circ$) from 1 view (0$^\circ$) on Case 1 from the SinoCT dataset, the input (red) and novel (green) angles are also indicated in the inset polar graph. We provide zoomed-in details for results with plausible anatomy compared to their corresponding ground truths.}
    \label{fig:s1-1}
\end{figure*}
\begin{figure*}[t]
    \centering
    \includegraphics[width=1\linewidth]{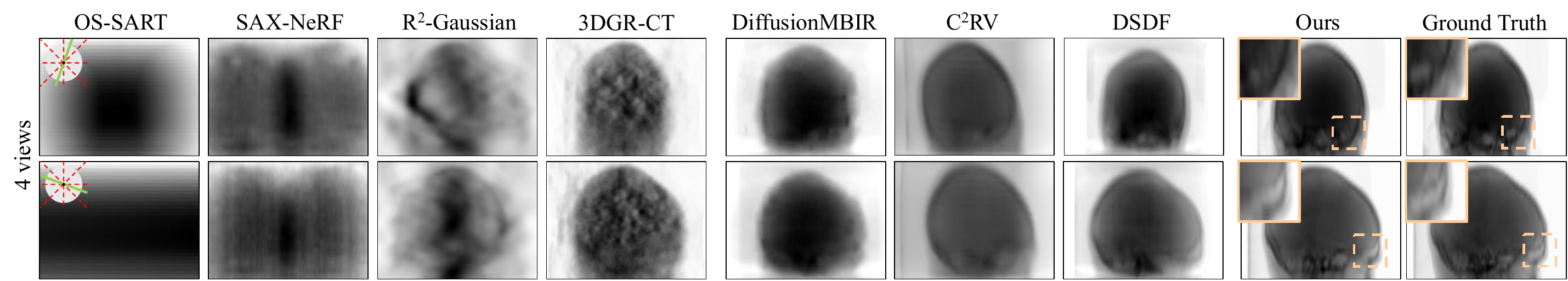}
    \caption{Novel View Synthesis results (22.5$^\circ$, 112.5$^\circ$) from 4 views (0$^\circ$, 45$^\circ$, 90$^\circ$, 135$^\circ$) on Case 1 from the SinoCT dataset, the input (red) and novel (green) angles are also indicated in the inset polar graph. We provide zoomed-in details for results with plausible anatomy compared to their corresponding ground truths.}
    \label{fig:s1-4}
\end{figure*}
\begin{figure*}[t]
    \centering
    \includegraphics[width=1\linewidth]{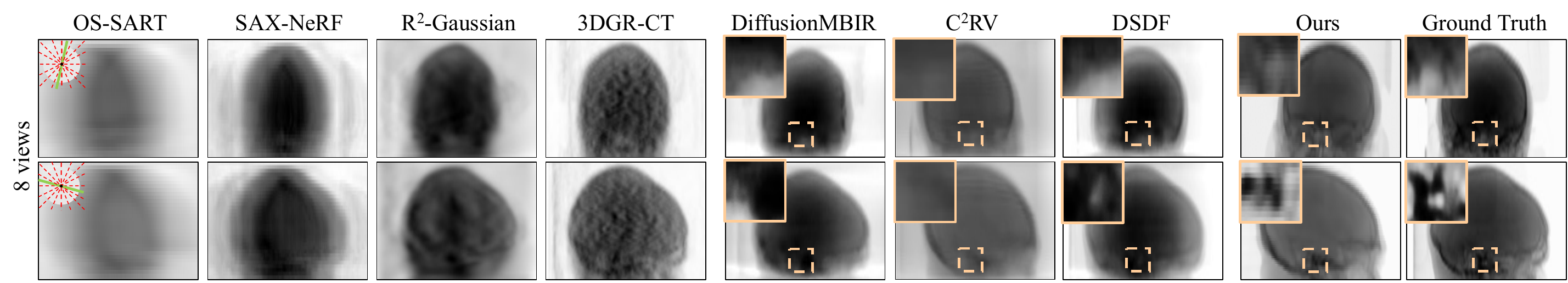}
    \caption{Novel View Synthesis results (11.25$^\circ$, 101.25$^\circ$) from 8 views (0$^\circ$, 22.5$^\circ$, 45$^\circ$, 67.5$^\circ$, 90$^\circ$, 112.5$^\circ$, 135$^\circ$, 157.5$^\circ$) on Case 1 from the CT-RATE dataset, the input (red) and novel (green) angles are also indicated in the inset polar graph. We provide zoomed-in details for results with plausible anatomy compared to their corresponding ground truths.}
    \label{fig:s1-8}
\end{figure*}
\begin{figure*}[t]
    \centering
    \includegraphics[width=1\linewidth]{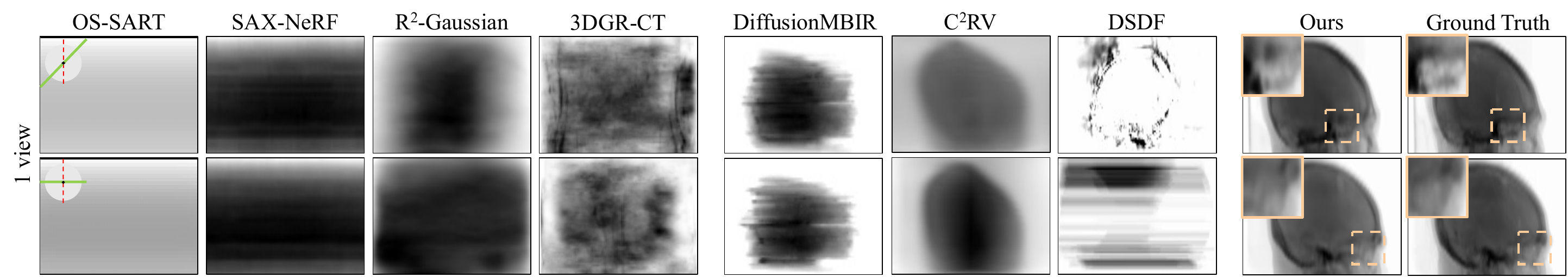}
    \caption{Novel View Synthesis results (45$^\circ$, 90$^\circ$) from 1 view (0$^\circ$) on Case 2 from the SinoCT dataset, the input (red) and novel (green) angles are also indicated in the inset polar graph. We provide zoomed-in details for results with plausible anatomy compared to their corresponding ground truths.}
    \label{fig:s2-1}
\end{figure*}
\begin{figure*}[t]
    \centering
    \includegraphics[width=1\linewidth]{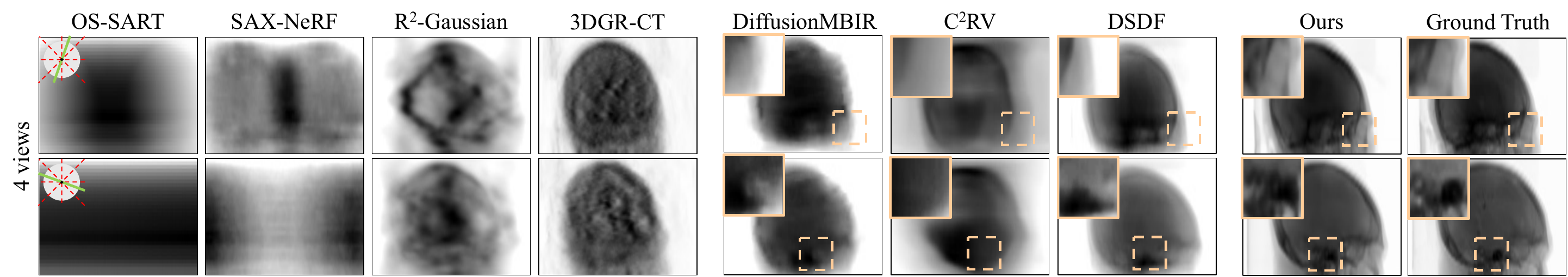}
    \caption{Novel View Synthesis results (22.5$^\circ$, 112.5$^\circ$) from 4 views (0$^\circ$, 45$^\circ$, 90$^\circ$, 135$^\circ$) on Case 2 from the SinoCT dataset, the input (red) and novel (green) angles are also indicated in the inset polar graph. We provide zoomed-in details for results with plausible anatomy compared to their corresponding ground truths.}
    \label{fig:s2-4}
\end{figure*}
\begin{figure*}[t]
    \centering
    \includegraphics[width=1\linewidth]{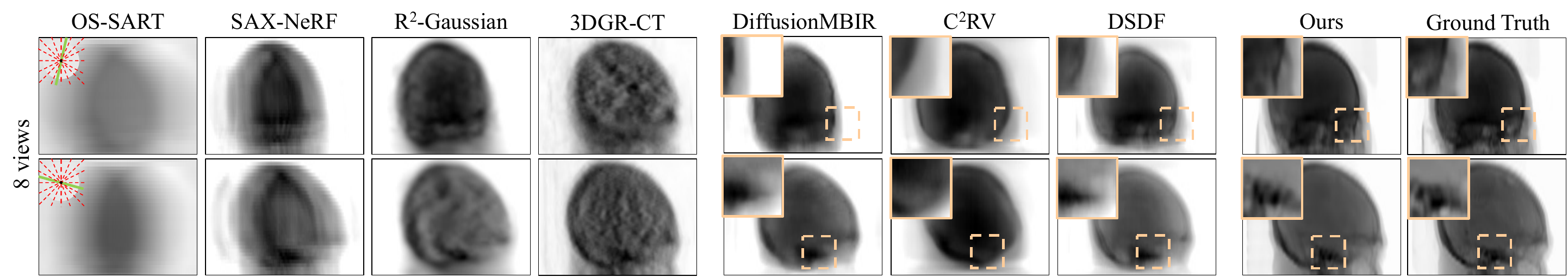}
    \caption{Novel View Synthesis results (11.25$^\circ$, 101.25$^\circ$) from 8 views (0$^\circ$, 22.5$^\circ$, 45$^\circ$, 67.5$^\circ$, 90$^\circ$, 112.5$^\circ$, 135$^\circ$, 157.5$^\circ$) on Case 2 from the SinoCT dataset, the input (red) and novel (green) angles are also indicated in the inset polar graph. We provide zoomed-in details for results with plausible anatomy compared to their corresponding ground truths.}
    \label{fig:s2-8}
\end{figure*}
\begin{figure*}[t]
    \centering
    \includegraphics[width=1\linewidth]{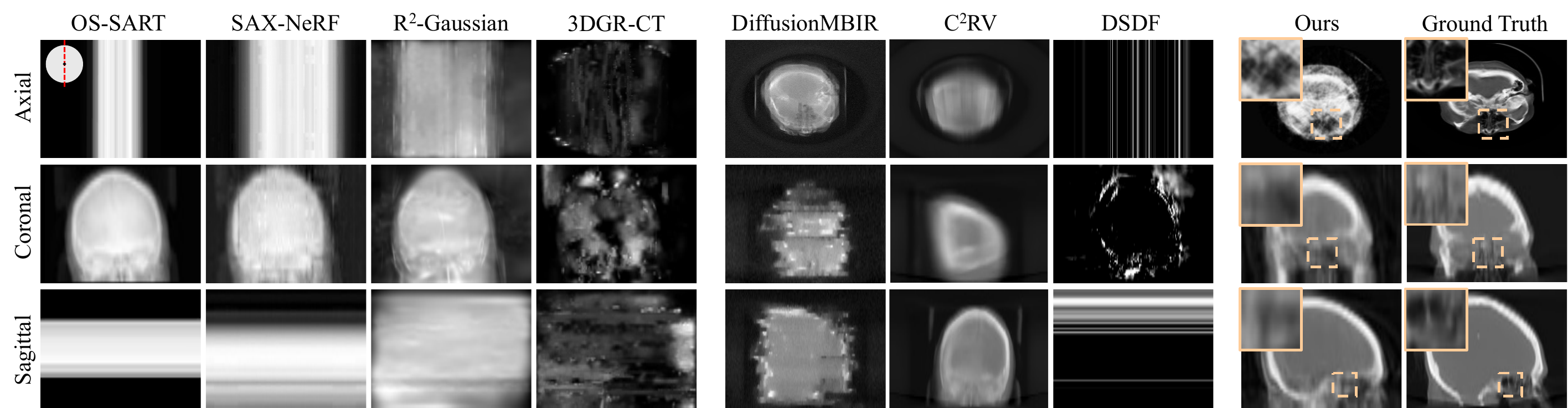}
    \caption{Tomographic Reconstruction results from 1 view (0$^\circ$) on Case 1 from the SinoCT dataset, the input angle is also indicated in the inset polar graph. We zoom in on methods with plausible details with the ground truths.}
    \label{fig:s1-1-}
\end{figure*}
\begin{figure*}[t]
    \centering
    \includegraphics[width=1\linewidth]{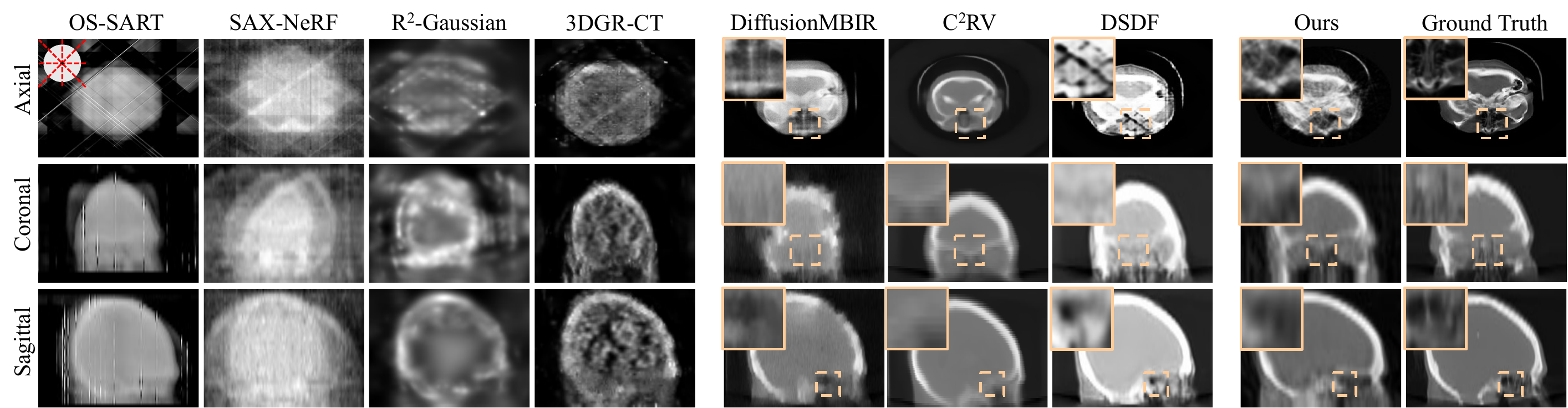}
    \caption{Tomographic Reconstruction results from 4 views (0$^\circ$, 45$^\circ$, 90$^\circ$, 135$^\circ$) on Case 1 from the SinoCT dataset, the input angles is also indicated in the inset polar graph. We zoom in on methods with plausible details with the ground truths.}
    \label{fig:s1-4-}
\end{figure*}
\begin{figure*}[t]
    \centering
    \includegraphics[width=1\linewidth]{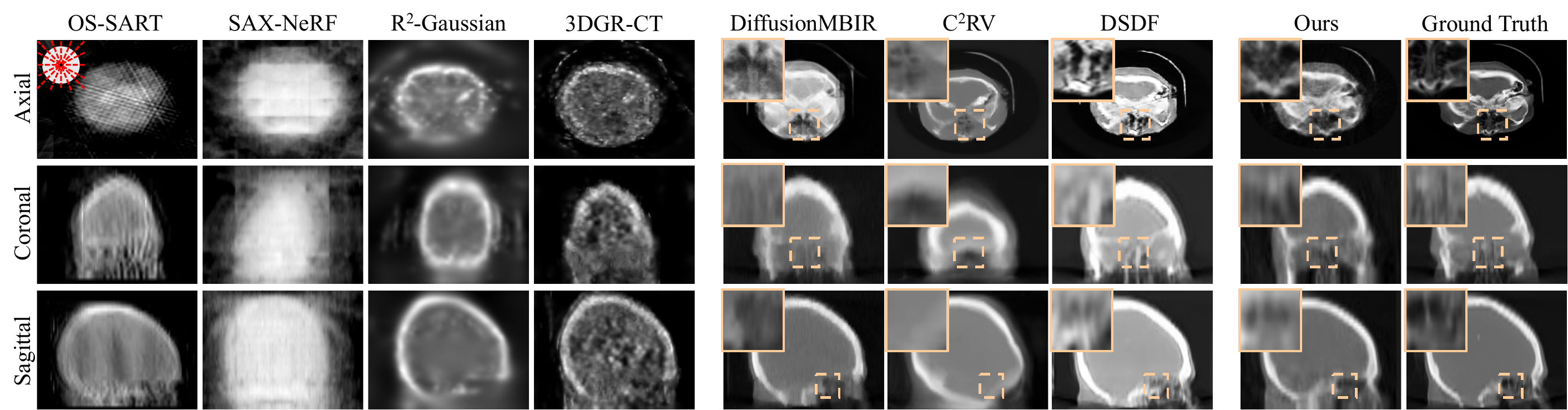}
    \caption{Tomographic Reconstruction results from 8 views (0$^\circ$, 22.5$^\circ$, 45$^\circ$, 67.5$^\circ$, 90$^\circ$, 112.5$^\circ$, 135$^\circ$, 157.5$^\circ$) on Case 1 from the SinoCT dataset, the input angles is also indicated in the inset polar graph. We zoom in on methods with plausible details with the ground truths.}
    \label{fig:s1-8-}
\end{figure*}
\begin{figure*}[t]
    \centering
    \includegraphics[width=1\linewidth]{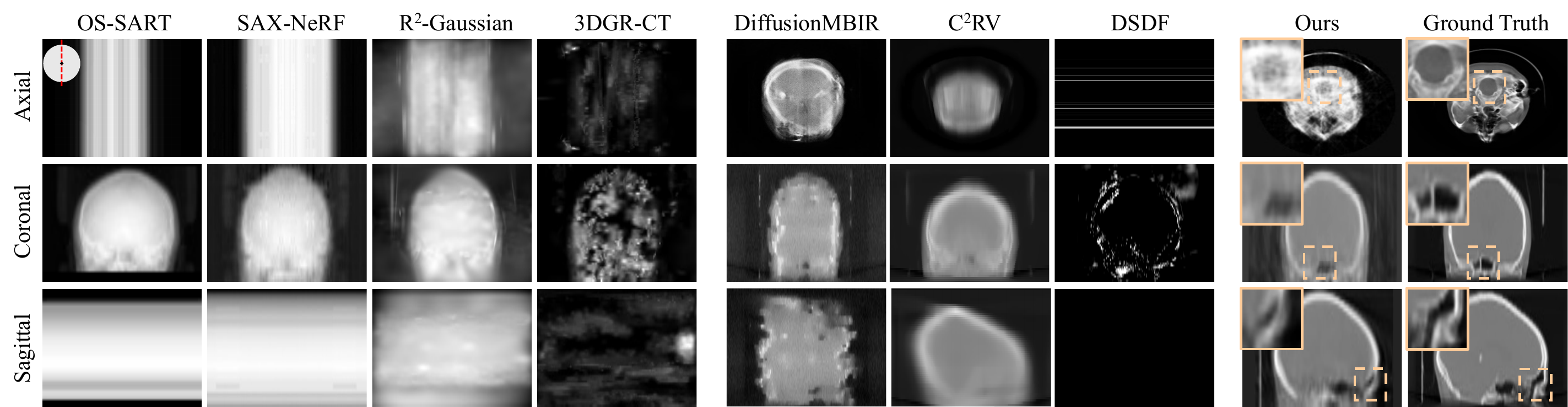}
    \caption{Tomographic Reconstruction results from 1 view (0$^\circ$) on Case 2 from the SinoCT dataset, the input angle is also indicated in the inset polar graph. We zoom in on methods with plausible details with the ground truths.}
    \label{fig:s2-1-}
\end{figure*}
\begin{figure*}[t]
    \centering
    \includegraphics[width=1\linewidth]{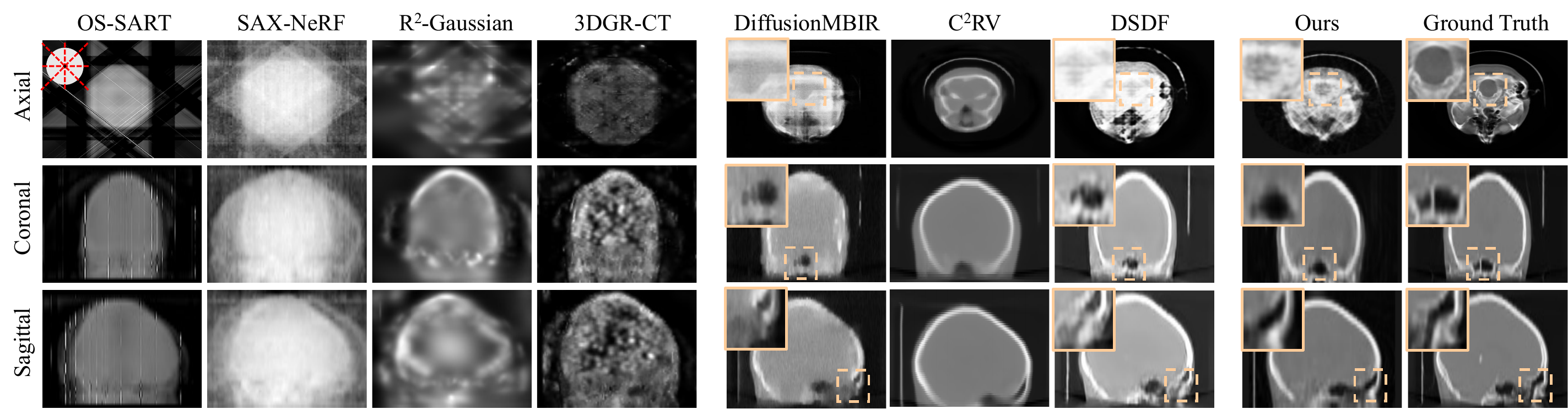}
    \caption{Tomographic Reconstruction results from 4 views (0$^\circ$, 45$^\circ$, 90$^\circ$, 135$^\circ$) on Case 2 from the SinoCT dataset, the input angles is also indicated in the inset polar graph. We zoom in on methods with plausible details with the ground truths.}
    \label{fig:s2-4-}
\end{figure*}
\begin{figure*}[t]
    \centering
    \includegraphics[width=1\linewidth]{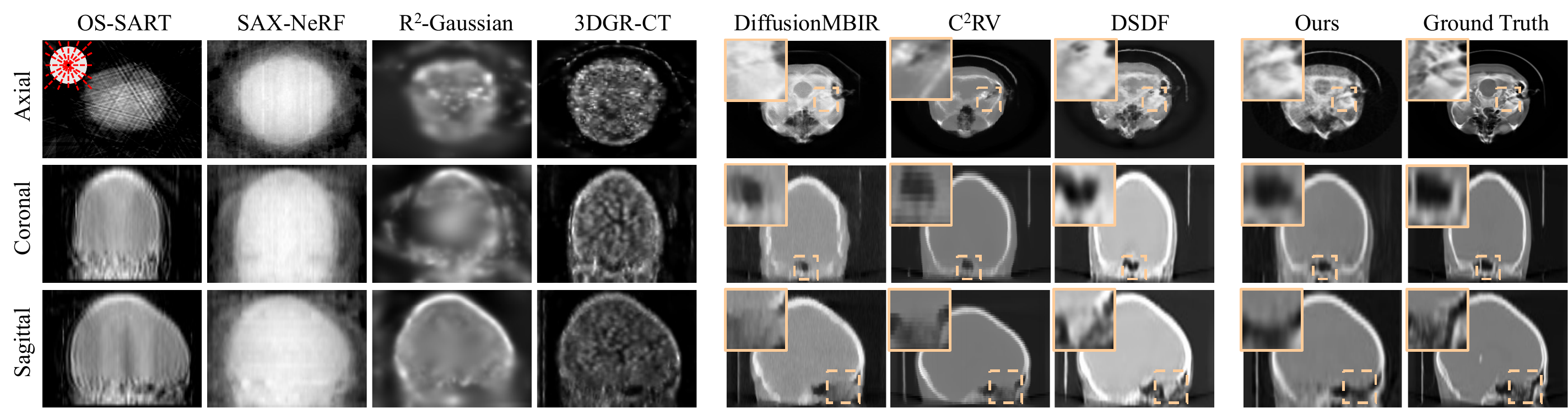}
    \caption{Tomographic Reconstruction results from 8 views (0$^\circ$, 22.5$^\circ$, 45$^\circ$, 67.5$^\circ$, 90$^\circ$, 112.5$^\circ$, 135$^\circ$, 157.5$^\circ$) on Case 2 from the SinoCT dataset, the input angles is also indicated in the inset polar graph. We zoom in on methods with plausible details with the ground truths.}
    \label{fig:s2-8-}
\end{figure*}
\clearpage

\section*{NeurIPS Paper Checklist}

\begin{enumerate}

\item {\bf Claims}
    \item[] Question: Do the main claims made in the abstract and introduction accurately reflect the paper's contributions and scope?
    \item[] Answer: \answerYes{} % Replace by \answerYes{}, \answerNo{}, or \answerNA{}.
    \item[] Justification: The main contribution of our paper is to enable the xSV-CT regime that requires extreme projection sparsity at both dataset curation and downstream application, which has been discussed in the abstract and introduction.

\item {\bf Limitations}
    \item[] Question: Does the paper discuss the limitations of the work performed by the authors?
    \item[] Answer: \answerYes{} % Replace by \answerYes{}, \answerNo{}, or \answerNA{}.
    \item[] Justification: The limitations of prior works has been discussed in abstract, introduction, and related works in more details.

\item {\bf Theory assumptions and proofs}
    \item[] Question: For each theoretical result, does the paper provide the full set of assumptions and a complete (and correct) proof?
    \item[] Answer: \answerNA{} % Replace by \answerYes{}, \answerNo{}, or \answerNA{}.
    \item[] Justification: This paper provides no novel theoretical results.

    \item {\bf Experimental result reproducibility}
    \item[] Question: Does the paper fully disclose all the information needed to reproduce the main experimental results of the paper to the extent that it affects the main claims and/or conclusions of the paper (regardless of whether the code and data are provided or not)?
    \item[] Answer: \answerYes{} % Replace by \answerYes{}, \answerNo{}, or \answerNA{}.
    \item[] Justification: The code, dataset links to public datasets used in this paper as well as data splits, and training checkpoints will be available in the github link in Appendix~\ref{sec:A}. Also, we provided extensive details of our dataset and implementation inside our appendices which helps reproduction.

\item {\bf Open access to data and code}
    \item[] Question: Does the paper provide open access to the data and code, with sufficient instructions to faithfully reproduce the main experimental results, as described in supplemental material?
    \item[] Answer: \answerNo{} % Replace by \answerYes{}, \answerNo{}, or \answerNA{}.
    \item[] Justification: We have disclosed our core algorithm with a pseudo code in the appendix, the full disclosure of code, dataset links, as well as model checkpoints will be available in the github link in Appendix~\ref{sec:A}.

\item {\bf Experimental setting/details}
    \item[] Question: Does the paper specify all the training and test details (e.g., data splits, hyperparameters, how they were chosen, type of optimizer) necessary to understand the results?
    \item[] Answer: \answerYes{} % Replace by \answerYes{}, \answerNo{}, or \answerNA{}.
    \item[] Justification: We include inference details are discussed in full in the Experiments section, exact splits along with other materials will be available in the github link in Appendix~\ref{sec:A}.

\item {\bf Experiment statistical significance}
    \item[] Question: Does the paper report error bars suitably and correctly defined or other appropriate information about the statistical significance of the experiments?
    \item[] Answer: \answerYes{} % Replace by \answerYes{}, \answerNo{}, or \answerNA{}.
    \item[] Justification: Standard deviation of $\ell_2$ error of our method across 4 runs on novel view synthesis are reported in the Experiments section as well as more are reported in the appendices.

\item {\bf Experiments compute resources}
    \item[] Question: For each experiment, does the paper provide sufficient information on the computer resources (type of compute workers, memory, time of execution) needed to reproduce the experiments?
    \item[] Answer: \answerYes{} % Replace by \answerYes{}, \answerNo{}, or \answerNA{}.
    \item[] Justification: The memory footprints and execution time can be found in the appendix.
    
\item {\bf Code of ethics}
    \item[] Question: Does the research conducted in the paper conform, in every respect, with the NeurIPS Code of Ethics \url{https://neurips.cc/public/EthicsGuidelines}?
    \item[] Answer: \answerYes{} % Replace by \answerYes{}, \answerNo{}, or \answerNA{}.
    \item[] Justification: We have read and followed the NeurIPS Code of Ethics.

\item {\bf Broader impacts}
    \item[] Question: Does the paper discuss both potential positive societal impacts and negative societal impacts of the work performed?
    \item[] Answer: \answerYes{} % Replace by \answerYes{}, \answerNo{}, or \answerNA{}.
    \item[] Justification: We have discussed the method's broader impacts in the appendices.
    
\item {\bf Safeguards}
    \item[] Question: Does the paper describe safeguards that have been put in place for responsible release of data or models that have a high risk for misuse (e.g., pre-trained language models, image generators, or scraped datasets)?
    \item[] Answer: \answerNA{} % Replace by \answerYes{}, \answerNo{}, or \answerNA{}.
    \item[] Justification: Our model is strictly trained on licensed public dataset sources, thus should not pose such risks.

\item {\bf Licenses for existing assets}
    \item[] Question: Are the creators or original owners of assets (e.g., code, data, models), used in the paper, properly credited and are the license and terms of use explicitly mentioned and properly respected?
    \item[] Answer: \answerYes{} % Replace by \answerYes{}, \answerNo{}, or \answerNA{}.
    \item[] Justification: We have read and complied with their dataset or code licenses, adequate citations have also been added to the main text.

\item {\bf New assets}
    \item[] Question: Are new assets introduced in the paper well documented and is the documentation provided alongside the assets?
    \item[] Answer: \answerYes{} % Replace by \answerYes{}, \answerNo{}, or \answerNA{}.
    \item[] Justification: We will provide README files with the codebase in the github link in Appendix~\ref{sec:A}.

\item {\bf Crowdsourcing and research with human subjects}
    \item[] Question: For crowdsourcing experiments and research with human subjects, does the paper include the full text of instructions given to participants and screenshots, if applicable, as well as details about compensation (if any)? 
    \item[] Answer: \answerNA{} % Replace by \answerYes{}, \answerNo{}, or \answerNA{}.
    \item[] Justification: This paper provides no such experiments.

\item {\bf Institutional review board (IRB) approvals or equivalent for research with human subjects}
    \item[] Question: Does the paper describe potential risks incurred by study participants, whether such risks were disclosed to the subjects, and whether Institutional Review Board (IRB) approvals (or an equivalent approval/review based on the requirements of your country or institution) were obtained?
    \item[] Answer: \answerNA{} % Replace by \answerYes{}, \answerNo{}, or \answerNA{}.
    \item[] Justification: This paper includes no such experiments.

\item {\bf Declaration of LLM usage}
    \item[] Question: Does the paper describe the usage of LLMs if it is an important, original, or non-standard component of the core methods in this research? Note that if the LLM is used only for writing, editing, or formatting purposes and does \emph{not} impact the core methodology, scientific rigor, or originality of the research, declaration is not required.
    %this research? 
    \item[] Answer: \answerNA{} % Replace by \answerYes{}, \answerNo{}, or \answerNA{}.
    \item[] Justification: LLMs were used strictly for paper polishing, and the core concept and code are originally constructed.

\end{enumerate}
\end{document}